\documentclass{article}
\pdfoutput=1
\usepackage{arxiv}

\usepackage[utf8]{inputenc} 
\usepackage[T1]{fontenc}    
\usepackage{hyperref}       
\usepackage{url}            
\usepackage{booktabs}       
\usepackage{amsfonts}       
\usepackage{amsmath}
\newcommand\numberthis{\addtocounter{equation}{1}\tag{\theequation}}

\usepackage{nicefrac}       
\usepackage{microtype}      
\usepackage{lipsum}
\usepackage{graphicx}
\usepackage{float}
\usepackage[natbibapa]{apacite}
\usepackage{algpseudocode,algorithm,algorithmicx}

\usepackage{graphicx}
\usepackage{subcaption}
\usepackage{float}

\newcommand{\KL}{D_\mathrm{KL}}
\newcommand{\E}{\mathbf{E}}

\begin{document}
\title{A Mathematical Walkthrough and Discussion of the Free Energy Principle}

\author{
 Beren Millidge \\
  School of Informatics\\
  University of Edinburgh\\
  \texttt{beren@millidge.name}
  \And
  \And
  Anil K Seth \\
  Sackler Center for Consciousness Science\\
   Evolutionary and Adaptive Systems Research Group\\
  School of Engineering and Informatics\\
  University of Sussex\\
  \texttt{A.K.Seth@sussex.ac.uk} \\
  \And
    Christopher L Buckley \\
  Evolutionary and Adaptive Systems Research Group\\
  School of Engineering and Informatics\\
  University of Sussex\\
  \texttt{C.L.Buckley@sussex.ac.uk} 
  
}

\maketitle   

\begin{abstract}
The Free-Energy-Principle (FEP) is an influential and controversial theory which postulates a deep connection between the stochastic thermodynamics of self-organization and learning through variational inference. Specifically, it claims that any self-organizing system which can be statistically separated from its environment, and which maintains itself at a non-equilibrium steady state, can be construed as minimizing an information-theoretic functional -- the variational free energy -- and thus performing variational Bayesian inference to infer the hidden state of its environment. This principle has also been applied extensively in neuroscience, and is beginning to make inroads in machine learning by spurring the construction of novel algorithms by which action, perception, and learning can all be unified under a single objective. While its expansive and often grandiose claims have spurred significant debates in both philosophy and theoretical neuroscience, the mathematical depth and lack of accessible introductions and tutorials for the core claims of the theory have often made productive discussion challenging. Here, we aim to provide a mathematically detailed, yet intuitive walk-through of the formulation and central claims of the FEP while also providing a discussion of the assumptions necessary and potential limitations of the theory. Additionally, since the FEP is a living theory, subject to internal controversy, change, and revision, we also present a detailed appendix highlighting and condensing current perspectives as well as controversies about the nature, applicability, and the mathematical assumptions and formalisms underlying the FEP.
\end{abstract}

\section{Introduction}
The Free-energy principle (FEP) is a grand theory, arising out of theoretical neuroscience, with deep ambitions to provide a unified understanding of the nature of self organisation under the rubric of Bayesian inference \citep{friston2006free,friston_free_2019,friston2010free,friston2012free}. Perhaps the central postulate of this theory is the `Free Energy Lemma' which states that one can interpret the internal dynamics of any self organizing system with a Markov Blanket (to be defined later), of any type and on any scale, as performing, on average, a kind of elemental Bayesian inference upon the external environment that surrounds it \citep{friston2013life,friston2012ao,friston2019particularphysics}. More generally, it claims to provide a recipe, in terms of the statistical dependencies which underpin a Markov Blanket, to define precisely and mathematically what it means to be a system at all \citep{friston2019particularphysics}. Understanding self-organization through the lens of inference provides a powerful perspective for understanding the nature of self-organizing systems, as it allows one to immediately grasp the nature of the dynamics which undergird self-organization, as well as apply the extremely large and powerful literature on Bayesian inference methods and algorithms to understanding the dynamics of self-organizing systems \citep{parr2020modules,parr2020markov,yedidia2011message}. 
Moreover, by framing everything in statistical terms -- in terms of conditional independence, generative models, and approximate posterior distributions -- the free-energy principle provides a novel and powerful vocabulary to talk about such systems, as well as to ask questions such as `what kind of generative model does this system embody?'. Ultimately, this new statistical and inferential perspective upon dynamics may lead to significant advances in our understanding of complex systems, both biotic and abiotic.

Historically, this perspective has close relationships with early cybernetic views of control and regulation \citep{wiener2019cybernetics,conant1970every,kalman1960new}. Philosophically, the FEP can be seen as an extension to Ashby's notion that every good regulator of a system must also become a model of the system \citep{conant1970every}. The FEP nuances this notion slightly by instead stating that every system that regulates itself against the external environment, must in some sense embody a generative model of the environment, and also that the dynamics of the internal states of the system necessarily can be interpreted as performing approximate variational inference upon a distribution over external variables of their local environment - in other words, such systems appear to form `beliefs' about the external sources of incoming fluctuations.

The free-energy principle originated in theoretical neuroscience, as an attempt to understand the mathematical properties that a self-organising living, biotic system, \emph{must} possess in order to sustain itself against thermodynamic equilibrium. It was first and especially applied to understanding the function of the brain \citep{friston2006free,friston2010action,friston2012history}, and has been developed into two main process theories -- predictive coding \footnote{Predictive coding as a theory has a long history, originating as a method for time-series compression \citep{spratling2017review} then being applied to processing in the brain first by \citet{srinivasan1982predictive} and \citet{mumford1992computational} and was related formally to variational inference and the free energy principle by \citep{friston2003learning}. For a recent review of predictive coding see \citep{millidge2021predictive}.} \citep{rao1999predictive,friston2003learning,friston2005theory,friston2008hierarchical} and active inference \citep{friston2009reinforcement,friston2012active,friston2015active,friston2017process,friston2018deep,da2020active} which have been investigated in a wide variety of paradigms, where it has been used to investigate a wide variety of phenomena from \citep{friston2014anatomy,friston2015knowing,friston2015active}, information foraging and saccades \citep{parr2017uncertainty,parr2018active,parr2019computational} exploratory behaviour \citep{schwartenbeck2013exploration,friston2015active,friston2017curiosity,friston2020sophisticated}, concept learning \citep{schwartenbeck_computational_2019}, and a variety of neuropsychiatric disorders \citep{lawson2014aberrant,benrimoh2019hallucinations,cullen2018active}. These process theories translate the abstract formulation of the FEP into concrete and practical algorithms by specifying certain generative models, variational distributions, and inference procedures, and have been shown to be useful both in providing  biologically plausible theories of learning and inference in the brain, and also in developing highly effective inference algorithms which have advanced the state of the art in machine learning \citep{parr2019neuronal,millidge_deep_2019,tschantz2020reinforcement,millidge2020relationship}. 

The FEP is (in)famously mathematically deep and broad, incorporating concepts and techniques from a wide range of disciplines such as advanced statistical methods for Bayesian inference, stochastic thermodynamics, classical physics, and differential geometry. The level of mathematical sophistication required, as well as sometimes dense expositions of the theory in key works, has made full comprehension of the core arguments and central results of the theory difficult to achieve, and has often resulted in confusion in the literature. In this tutorial, we aim to provide a self-contained and intuitive walk-through of the key mathematical results underpinning the FEP, relying on a fairly minimal amount of prerequisite knowledge -- specifically, linear algebra, probability theory and statistics, and basic concepts of differential equations. We also aim to derive all results in a fully explicit way, with substantial commentary to aid intuitive understanding and to make clear all assumptions and major logical moves in the argument. Finally, we provide a fairly detailed discussion on the nature and meaning of the core results of the FEP as well as the potential limitations of the theory.

Additionally, we also offer a substantive appendix which includes in-depth discussions of all the assumptions and potential limitations highlighted in the main text, as well as current theoretical debates within the free energy community over the mathematical formalism and general applicability of the FEP. Thus, the main text presents the FEP in its best possible light, while the potential controversial assumptions and other mathematical difficulties, which are substantial and still under significant discussion within the community, are presented in the footnotes to the main text and then discussed in detail in the appendix. As such, critical readers should make sure to consult the appendix in detail for a balanced understanding of both the claims, and the potential limitations of the free energy principle

\subsection{Related Work}

While there have been many excellent tutorials on the process theories arising from the free-energy principle, such as predictive coding \citep{bogacz2017tutorial,buckley2017free, millidge2021predictive}, and active inference \citep{da2020active,friston2017process} , there is a general lack in the literature of mathematically detailed tutorials tackling the central claims of the FEP, especially in its most recent formulation in the \citet{friston2019particularphysics}. This monograph currently represents the theoretical state of the art of the Free-Energy-Principle \footnote{Many of the key results are also reprised more concisely in \citet{parr2020markov,da2021bayesian}}, and we aim to recapitulate its main results in a more direct and accessible manner in this paper.

The FEP also involves deep knowledge from a variety of disciplines. For thorough tutorials on variational Bayesian inference, we recommend \citep{fox2012tutorial,beal2003variational,blei2017variational}. For an excellent walk through of differential geometry, information-geometry, and the role of the Fisher Information matrix, we recommend \citet{caticha2015basics}. For a detailed treatment of the principles of stochastic thermodynamics which underlie much of the FEP, we recommend \citet{seifert2008stochastic,seifert2012stochastic,esposito2010three1,van2010three2}. Finally, since the publication of the particular physics monograph \citep{friston2019particularphysics}, the theory of the FEP has been developed in numerous further publications. A concise overview of the theory is given in \citet{parr2020markov}, and later developments have been surveyed in \citet{da2021bayesian}. Further development of the operationalization of Markov Blankets is given in \citet{parr2020modules}. Discussion upon different inference algorithms and divergences has been given in \citet{yedidia2011message,blei2017variational}. Some further theoretical refinements to various aspects of the theory can be found in \citep{ramstead2020neural,friston2020sentience}.

For a detailed mathematical critique to many of the arguments and claims of the FEP, especially regarding the technical conditions required for the free energy lemma, please see \citet{biehl2020technical}, and also the response \citet{friston2020some}, as well as \citet{aguilera2021particular} for additional critiques of the markov blanket condition, assumptions around the solenoidal flows, and the some of the technical steps and assumptions of the free energy lemma. 

\subsection{History and Logical Structure}

\begin{figure}
    \centering
    \includegraphics[scale=0.2]{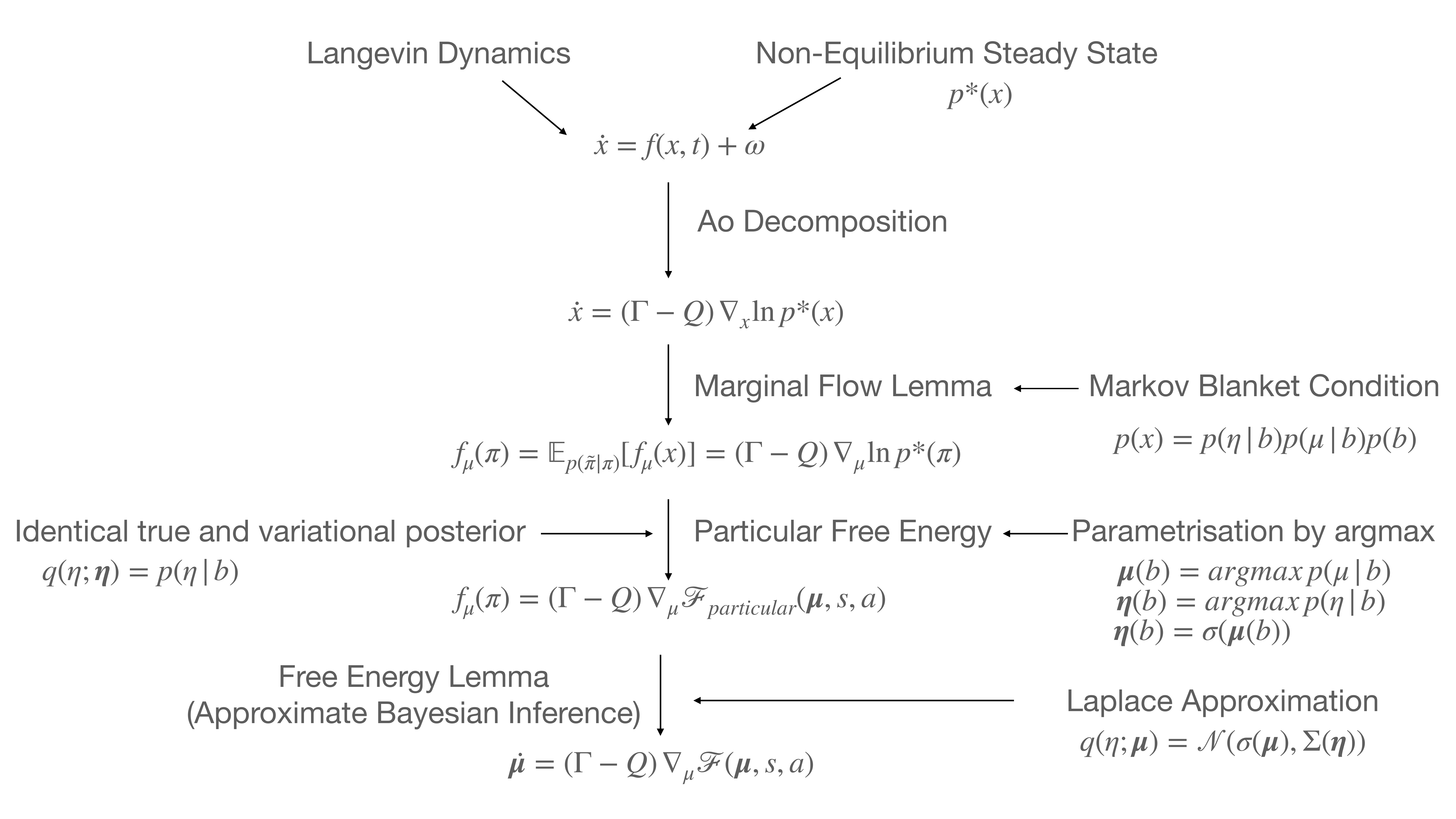}
    \caption{The logical flow of the argument of the FEP from the initial formulation to the crucial approximate Bayesian inference lemma. We begin with a setting of a random langevin stochastic dynamical system, which possess a non-equilibrium-steady state. By applying the Ao decomposition, we can understand its dynamics in terms of a gradient descent upon the surprisal. Upon the addition of a Markov Blanket partition, we can express subsets in terms of their own marginal flows via the marginal flow lemma. If we then identify the internal states as parametrizing a variational distribution over the external states, we can interpret the marginal flow on the surprisal as a flow on the variational free energy, under the Laplace approximation, and thus interpret the internal states of the system as a whole as performing a simple kind of variational inference upon the external states.}
    \label{logical_structure}
\end{figure}
Historically, the free energy principle has evolved over the course of about fifteen years. Its intellectual development can best be seen in two phases. In the first phase, an intuitive and heuristic treatment emerged with \citet{friston2006free} which stated that the imperative to minimize variational free energy emerged from a necessary imperative of minimizing the system's entropy, or log model evidence, which is upper bounded by variational free energy. This imperative emerges due to the self-sustaining nature of biological systems such as brains, in that they maintain a set distribution against the inexorably increasing entropic nature of thermodynamic reality \citep{friston2009free}. In order to do so, systems must constantly seek to reduce and maintain their entropy across their state space. Since the VFE is computationally tractable while the entropy itself is not, it was postulated that neural systems maintain themselves by implicitly minimizing this proxy rather than the actual entropy itself \citep{friston2010free}.

Later, in the second phase \citep{friston2013life}, this heuristic argument and intuition was related more formally to concepts in stochastic thermodynamics beginning with \citep{friston2012ao,friston2012free}. Specifically, the framework developed mathematically into a description of stochastic dynamics (as stochastic differential equations) separated into `external, internal, and blanket' states by a statistical construct called a \emph{Markov Blanket}. This blanket makes precise the statistical independence conditions required to make sense of talking about a `system' as distinct from its `environment'. Moreover, by separating the `blanket' into `sensory' and `active' states, one can obtain a statistical description of the core elements of a perception-action-loop, a central concept in cybernetics, control theory, and reinforcement learning. Secondly, the theory developed a precise notion of what it means to maintain a stable `phenotype' which is interpreted mathematically as a non-equilibrium steady-state density over the state-space. This steady state is non-equilibrium due to the presence of `solenoidal flows' which are flows orthogonal to the gradient of the NESS density. Mathematically, such flows do not increase or decrease the entropy of the steady-state-density, but do, however, in contrast to an equilibrium steady state (ESS), provide a clear arrow of time. Given this, it is claimed, that under certain conditions, one can draw a relationship between the flow dynamics and the process of variational Bayesian inference through the minimization of the variational free energy -- specifically, that the dynamics that result from this specific kind of flow under a Markov blanket at the NESS density can be seen as approximating a gradient descent upon the VFE, thus licensing the interpretation of the system as performing a basic kind of Bayesian inference or, `self-evidencing' \citep{hohwy2008predictive,clark2015surfing}

While the intuitions and basic logical structure of the theory has remained roughly constant since \citet{friston2013life}, the mathematical formulation and some of the arguments have been refined in the most recent \citet{friston2019particularphysics} monograph and related papers \citep{friston2020some,parr2020markov,da2021bayesian}. These papers have drawn close connections between the formulation of free energy principle, and many aspects of physics including the principle of least action in classical mechanics, and notions of information length and the arrow of time in stochastic thermodynamics. Additionally, recent work contains a novel information-geometric gloss on the nature of the Bayesian inference occurring in the system. Specifically, it argues that the internal states of the system can be seen as points on an \emph{statistical manifold} that parametrize distributions over the external states, and that thus the internal states can be described using a `dual-aspect information geometry.' According to this perspective, internal states evolve in both the `intrinsic' state space of the system's physical dynamics, while simultaneously parameterising a manifold of statistical beliefs about external states - the so-called `extrinsic' information geometry. 

While the mathematical depths of the FEP often appears formidably complex to the uninitiated, the actual logical structure of the theory is relatively straightforward. First, we want to define what it means to be `a system' that keeps itself apart from the outside `environment' over a period of time. The FEP answers this question in its own way. We define a `system' (according to the FEP) as a dynamical system which has a non-equilibrium steady state (NESS) which it maintains over an appreciable length of time, and that the dynamics are structured in such a way that they obey the `Markov Blanket Condition'. Specifically, having a NESS can be intuitively thought of as defining dynamics which produce something like a phenotype -- i.e. a recognizable pattern of states which persists relatively unchanged for some period of time. For instance, we can think of the biological systems in such a manner. Biological organisms maintain relatively steady states, against constant entropic dissipation, for relatively long (by thermodynamic standards) periods of time. Of course, from a purely thermodynamical perspective, in resisting entropy themselves, biological organisms are not countering the law of thermodynamics. To achieve their steady state requires a constant influx of energy -- hence it is a non-equilibrium steady state (NESS). From this perspective, we can understand biological organisation to be the process of creating `dissipative structures' \citep{prigogine1973theory, kondepudi2014modern} which only manage to maintain themselves at steady state and reduce their own entropy at the expense of consuming energy and increasing the entropy production rate of their environment \citep{prigogine2017non}. Illustrative physical examples of similar NESS states are Benard convection cells, and the Belousov-Zhabotinsky reaction \citep{zwanzig2001nonequilibrium}. In practical terms, we can consider the NESS density to be the `phenotype' of the system. From the perspective of the FEP, we are not usually concerned with whether a set of dynamics possesses a NESS density, or how convergence \emph{to} the NESS density works. Instead we take it as an axiom that we possess a system with a NESS density which the system can converge to, and are instead concerned with the dynamical behaviour of the system \emph{at} the NESS density. While this is clearly a special case, nevertheless dynamical systems at NESS already exhibit rich behaviours to effectively maintain themselves there. It is these properties, which any system which maintains itself at NESS must possess, that are the fundamental object of study for the FEP. 

Secondly, now that we have a set of dynamics which has a NESS density, and thus exhibits some stability through time, we also require a statistical way to separate the `system' from the `environment'. The FEP handles this by stipulating that any system it considers must fulfil a set of criteria which we call the Markov Blanket conditions. These conditions, deriving from the idea of Markov blankets in Bayesian networks \citep{pearl2011bayesian,pearl2014probabilistic}, set forth a set of conditional independence requirements that allow a system to be statistically separated from its environment \footnote{Whenever we say Markov Blanket, following standard use in the literature, we mean the \emph{minimal} Markov blanket -- i.e. the Markov Blanket which requires the fewest number of blanket states to achieve the required conditional independencies.}. Specifically, we require that the dynamics of the system can be meaningfully partitioned into three sets of states -- `internal' states which belong to the system of study, `external' states which correspond to the environment, and `blanket states' which correspond to the boundary between the system and its environment. Specifically, we require the internal states to be conditionally independent of the external states given the blanket states, and vice versa. Thus all `influence' of the environment must travel through the blanket, and cannot directly interact with the internal states of the system which are `shielded' behind the blanket \footnote{Interestingly, mathematically, the MB condition and all of the FEP is completely symmetrical between `internal' and `external' states. Thus from the perspective of the system, the `external states' are its environment, but from the perspective of the environment, the `external states' are the system. This means that the environment models and performs inference about the system just as the system models and performs inference on the environment. We can thus think of the environment-system interaction as a duality of inference, where each tries to model and infer the other in a loop.}

Now that we have a system with a NESS density which obeys the Markov Blanket conditions, so that we can partition it into external, internal, and blanket states, we then wish to understand the dynamics of the system \emph{at} the NESS density, so we can understand the necessary behaviours of the system to allow the NESS to be maintained. Here we use the Helmholtz (Ao) decomposition \citep{yuan2017sde,yuan2011potential,yuan2012beyond} to represent the dynamics as a gradient flow on the log of the NESS density (which is called the surprisal) with both dissipative (in the direction of the gradient) and solenoidal (orthogonal to the gradient) components. Now that we can express the dynamics of the system in terms of gradients of the log NESS density, we then invoke the \emph{Marginal Flow Lemma} to write out the dynamics of each component of the partitioned dynamics (i.e. external, internal, and blanket states) solely in terms of a gradient flow on its own marginal NESS density. This means that we can express, for instance, the dynamics of the internal states solely in terms of gradient flows on the marginal NESS density over the internal and blanket states.

Given this marginal partition, we can analyze and understand each of the flows in each partition of the system separately. Specifically, to understand the Ashbyan notion that `every good regulator of a system is a model of the system', we wish to understand the relationship between the flows of the internal and external states, which are statistically separated from the blanket. Despite this separation, it is possible to define a mapping between the most likely internal state, given a specific configuration of the blanket states, and the distribution over the most likely external state of the system. We can use this mapping to interpret internal states as parametrizing a variational or approximate distribution \emph{over} the external states. This interpretation sets up the `dual-aspect' information geometry of the internal states, since the internal dynamics simultaneously represent changes in the parameters of the distribution over internal states (which can potentially be non-parametric), and changes to the parameters of the variational distribution over external states. This latter interpretation means that the internal states can be directly mapped to parameters of a distribution over external states, and thus that these parameters form a statistical manifold equipped with a Fisher information metric (if the variational distribution is in the exponential family), and in general becomes amenable to the techniques of information geometry \citep{amari1995information,ollivier2017information} Finally, given that we can interpret the internal states as parametrising a \emph{distribution} over external states, we can reconsider the gradient flow upon the log NESS density with a new light. Specifically, once the identification of the dynamics of the internal mode to variational inference is recognised, we can understand the NESS density to represent the implicit \emph{generative model} of the system (since it is a joint density over all variables of the dynamical system modelled), and the gradient flow dynamics as a descent upon the free-energy, with a perfect Bayes-optimal posterior. Alternatively, if we invoke an approximate posterior distribution over external states, we can represent the gradient flow as an approximate minimization of the variational free energy (VFE), and thus the internal states of the system can be interpreted as performing approximate variational Bayes. This is the \emph{key result} of the FEP. It states, simply, that the internal dynamics of any system that maintains itself at a non-equilibrium steady state, and possesses a Markov Blanket, can be interpreted as modelling, and performing approximate variational inference upon the external states beyond its own Markov Blanket. It thus generalizes and makes precise Ashby's notion that every good regulator must in some sense be a model of the system \citep{conant1970every}. Here we see that in order to maintain a non-equilibrium steady state, to counteract the dissipative forces inherent in thermodynamics, it is necessary to perform some kind of inference about the environment beyond the system itself. 

\section{Formulation}

Here we begin the precise mathematical description of the FEP. We aim to provide a consistent notation, and more detailed derivations of key results than are often presented. The presentation in this tutorial mostly follows the order of presentation in \citet{friston2019particularphysics}, although many circumstantial topics are omitted to focus on the main flow of the argument. We begin with the basic mathematical setting and formulation of the theory. We assume that the dynamics we wish to describe can be expressed in terms of a Langevin stochastic differential equation \citep{jaswinskistochastic},
\begin{align*}
\label{SDE_Dynamics}
\frac{dx}{dt} = f(x) + \omega \numberthis
\end{align*}
where $x = [x_0 \dots x_N]$ is a vector of states of some dimensionality, and f(x) is an arbitrary nonlinear but differentiable function of the state vector. Specifically, here we assume already that this process is not history dependent. The dynamics only depend on the instantaneous values of the states. In practice, history dependent systems can be represented in this fashion, albeit somewhat unintuitively by adding sufficient statistics of the history to the state itself. $\omega$ is assumed to be white (zero autocorrelation) Gaussian noise with zero mean such that $\omega$ = $\mathcal{N}(x; 0, 2 \Gamma)$ where $\Gamma$ is the variance of the noise. Zero autocorrelation means that the covariance between the noise at any two time instants, even infintesimally close together, is 0  --  $\mathbf{E}[\omega_t \omega_{t+\delta}^T] = 0$. We assume that this noise is added additively to the dynamics. A full list of the assumptions required in the formulation are outlined and discussed in the Appendix (section 10).

This stochastic differential equation can also be represented not in terms of dynamically changing states, but in terms of a dynamically changing \emph{probability distribution} over states. This transformation is achieved through the Fokker-Planck equation, by which we can derive that the change in the distribution over states can be written as,
\begin{align*}
\label{Fokker_Planck}
\frac{d p(x,t)}{dt} = - \nabla_x f(x)p(x,t) + \Gamma \nabla_x^2 p(x,t) \numberthis
\end{align*}

Where $p(x,t)$ is the instantaneous distribution over the states at a given time $t$. $p(x,t)$ begins with the distribution $p(x_0, 0) = \mathcal{N}(f(x_0,0), \Gamma)$ due to the initial noise term $\omega$ and the starting state $x_0$. Here $\nabla_x f(x,t)$ is the gradient function and simply denotes the vector of partial derivatives of the function $f$ with respect to each element of the vector $x$. $\nabla_x f(x) = [ \frac{\partial f(x,t)}{\partial x_0}, \frac{\partial f(x,t)}{\partial x_N}, \dots , \frac{\partial f(x,t)}{\partial x_N}]$. $\nabla^2_x f(x)$ represents the vector of second partial derivatives of the function.

Next, we presuppose that the dynamics expressed in Equation \ref{SDE_Dynamics} tend towards a non-equilibrium steady state $\lim_{t \to \infty} p(x,t) = p^*(x)$ where we represent the steady state distribution as $p^*(x)$. Note that this distribution no longer depends on time, since it is by definition at a steady state. We use $p^*$ to make clear that this distribution is at steady state. By definition a steady state distribution does not change with time, so that $\frac{dp^*(x)}{dt} = 0$.

The distinction between an equilibrium steady state and a non-equilibrium steady state (NESS) distribution is subtle and important. An equilibrium steady state, mathematically, is one where the property of detailed balance holds. This means that any transition between states at equilibrium is just as likely to go in the `forwards' direction as it is to go in the `backwards' direction. In effect, the dynamics are completely symmetric to time, and thus there is no notion of an arrow of time in such systems. Conversely, a non-equilibrium steady state is one where detailed balance does not hold, so there is a directionality to the dynamics, and thus an arrow of time, even though the actual distribution over states remains constant. From a thermodynamic perspective, the equilibrium-steady-state is the inexorable endpoint of the second law of thermodynamics, since it is the maximum entropy state. Conversely, a NESS is not a maximum entropy solution, since the directionality of the dynamics means that there is a degree of predictability in the system which could in theory be exploited to produce work. Non-equilibrium steady states can arise in thermodynamic systems but require an external source of driving energy as a constant input to the system, which is then dissipated to the external surroundings and gives the NESS a positive entropy production rate. To take an intuitive example, we can think about the thermodynamic equilibrium of a cup of coffee with cream added. The equilibrium steady state (ESS) is when the coffee and cream have completely diffused into one another, so that the cream maintains a constant proportion throughout the entire coffee cup. This will be the inevitable result (by the second law of thermodynamics) of adding an initially low entropy highly concentrated cream scoop into the coffee. On the other hand, we can think of the non-equilibrium steady state (NESS) as to be when the cream and coffee are equally diffused throughout, but somebody is constantly stirring the coffee in a specific direction. Here, we are at steady state because the concentrations of cream and coffee don't change over time, but nevertheless there is a directionality to the dynamics in the direction of the stirring. This directionality is only maintained due to a constant input of energy \footnote{It's important to note that here we are using physical intuition and concepts like `energy' in a purely metaphorical sense. All results here apply to arbitrary SDEs which do not necessarily follow the same constraints as physical systems -- i.e. respect conservation of energy} to the system (the stirring) \footnote{Interestingly, physical experience with this analogy would suggest that the solenoidal dynamics leading to NESS would lead to faster convergence to the NESS density compared to the strictly dissipative dynamics leading to ESS -- effectively, stirring helps the cream diffuse faster.  This insight has been applied to the design of highly efficient Markov-Chain-Monte-Carlo samplers in machine learning \citep{metropolis1953equation,neal2011mcmc,betancourt2013generalizing}}. The flow caused by the stirring is referred to as the `solenoidal flow' and mathematically is necessarily orthogonal to the gradient of the steady state distribution. This is necessary so that the solenoidal flow does not ascend or descent the gradient of the density, and thus change the steady state distribution which, as a steady state, by definition cannot change. Biological self organizing systems are often considered to be `dissipative structures', or non-equilibrium steady states from the perspective of thermodynamics, since they maintain a relatively steady state over time which requires a constant influx of energy to maintain. 

Given that we presuppose a system with a NESS density, we wish to understand the dynamics \emph{at} the NESS density -- specifically, how does the solenoidal flow help prevent the system from relaxing into an equilibrium-steady-state (ESS)? To understand this, we utilize the Helmholtz decomposition \citep{yuan2017sde,yuan2012beyond,friston2012ao} to rewrite the dynamics at the NESS into a form of a dissipative and solenoidal descent upon the gradient of the log NESS density. The Helmhotlz (or Ao) decomposition is a matheamtical tool which lets us express the `flow' -- i.e. dynamics function $f$ -- of a dynamical system into separable `dissipative' (noise) and `solenoidal' components which perform a gradient descent on a scalar potential function which we identify with the NESS. 
Mathematically, the Helmholtz decomposition can be written as,
\begin{align*}
f(x) = (\Gamma(x) - Q(x))\nabla_x \ln p^*(x) \numberthis
\end{align*}
Where $\Gamma(x)$ is a dissipative component of the flow which tries to descend the log density. It is the amplitude of the random fluctuations in the original SDE formulation \citep{jordan1998variational,yuan2010constructive,yuan2011potential}, which in effect are constantly trying to `smooth out' the NESS density and increase its entropy. Conversely, the $Q(x)$ represents the solenoidal portion of the flow which, although orthogonal to the gradient of the log potential, successfully counteracts the dissipative effects of the $\Gamma(x)$ terms to maintain the dynamics at a steady state. While $\Gamma(x)$ and $Q(x)$ can in theory be state-dependent, from here on out we typically assume that they are not -- $\Gamma(x) = \Gamma$; $Q(x) = Q$ \footnote{Specifically, the introduction of state-dependent $\Gamma$ and $Q$ matrices leads to there being an additional `housekeeping' term in the Helmholtz free energy as discussed in \citep{friston2021stochastic}.} We verify that the Helmholtz decomposition is satisfied at steady state in Appendix \ref{helmholtz_decomp_appendix}.

\section{Markov Blankets}

From these preliminaries, we have a set of dynamics of states $x$, which possess a NESS density, and, by using the Ao decomposition, we can express the dynamics at the NESS density in terms of dissipative $\Gamma$ and a solenoidal $Q$ flows on the gradient of the log density. Now, we begin to explore the statistical structure of these dynamics in terms of a Markov Blanket. Specifically, we next require that we can partition the states $x$ of the dynamics into three separate units. External states $e$, internal states $i$, and blanket states $b$ such that $x = [e,i,b]$. Intuitively, the external states represent the `environment'; the internal states represent the `system' we wish to describe, and the blanket states represent the statistical barrier between the system and its environment. For instance, we might wish to describe the dynamical evolution of a simple biological system such as a bacterium in such a manner. Here, the internal states would describe the internal cellular environment of the bacterium -- the cytoplasm, the nucleus, the ribosomes etc. The external states would be the environment outside the bacterium, while the blanket states would represent the cell membrane, sensory epithelia, and potentially active instruments such as the flagella which sense and interact with the external environment. The key intuition behind the FEP is that although all influence between external and internal states is mediated by the blanket states, simply maintaining the non-equilibrium steady state against environmental perturbations requires that the internal states in some sense model and perform (variational) Bayesian inference on the external states. The Markov Blanket condition is straightforward. It simply states that the internal and external states must be independent given the blanket states,
\begin{align*}
\label{Markov_Blanket_Independencies}
p^*(x) = p^*(e,i,b) = p^*(e | b)p^*(i|b)p^*(b) \numberthis
\end{align*}

\begin{figure}
    \centering
    \includegraphics[scale=1]{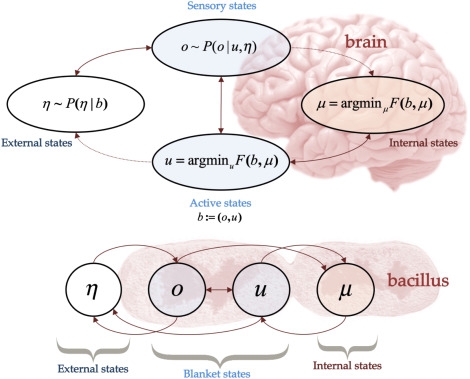}
    \caption{The intuition behind the Markov Blanket partition. The brain (or bacillus) consists of internal states $\mu$ which are separated from the outside world (external states $\eta$ by the blanket states $b$, which can themselves be partitioned into sensory states $s$, representing the sensory epithelia, and which are directly influenced by external states, and active states $a$ representing the organisms effectors and which are directly influenced by internal states, and act on external states. We see that perception concerns the minimization of free energy of the \emph{internal states}, while action concerns the minimization of the expected free energy of the \emph{active states}. Figure originally appeared in \citet{friston2019particularphysics}}
    \label{fig:my_label}
\end{figure}

While in probabilistic terms this factorisation is straightforward, it has more complex consequences for the dynamical flow of the system. Firstly, we additionally decompose the blanket states into sensory $s$ and active $a$ states such that $b = [s,a]$ and thus, ultimately $x = [e,i,s,a]$. Sensory states are blanket states that are causal children of the external states -- i.e. the states that the environment acts on directly. Active states are those blanket states that are not causal children of the external states. Essentially, external states influence sensory states, which influence internal states, which influence active states, which influence external states. The circular causality implicit in this loop is what allows the Markov Blanket condition to represent the perception-action loop \footnote{There has been some recent controversy in the literature about the meaning and implications of the statistical Markov blanket condition in terms of what it does or does not imply about the dynamical `real' connectivity. For more discussion on this please see the Appendix section 10.3.1 or \citet{aguilera2021particular}}. 

The next step is to understand what the conditional independence requirements put forth in Equation \ref{Markov_Blanket_Independencies} imply for the dynamics of the flow. Specifically, we obtain the \emph{marginal flow lemma} (see \citet{friston2019particularphysics} for a full derivation), which states that the flow of a subset of variables averaged under the rest of the variables, is equal to a solenoidal gradient flow only on the gradient of the potential of \emph{marginal NESS density} of that subset of variables plus a solenoidal coupling term. For instance, we can express the flow of just the external states $e$, averaged under the blanket and internal states, as simply a solenoidal gradient descent flow on the marginal NESS densities plus a term dependent upon the solenoidal coupling of the external and blanket states

\begin{align*}
    f_e(x) = \mathbb{E}_{p(e | b)}\big[ f_e(x)] = (\Gamma_{e,e} - Q_{e,e}) \nabla_e \ln p^*(e) - Q_{e,b} \nabla_b p(e,b)
\end{align*}

If we then assume a lack of solenoidal coupling so that $Q_{e,b} = 0$ then we can neglect this  solenoidal coupling term and treat the marginal flow as completely independent gradient descent. This lets us investigate in detail the information-theoretic interactions of one set of states with another, and gain intuition and understanding of the core information-theoretic properties of the perception-action loop. Similarly, using the marginal flow lemma we can express the flow of autonomous (active and internal) $\alpha = (a,i)$ as,
\begin{align*}
f_\alpha(x)= (\Gamma_{\alpha \alpha} - Q_{\alpha \alpha}) \nabla_\alpha \ln p^*(i,s,a) \numberthis
\end{align*}
So we see that autonomous states follow a gradient descent on the marginal NESS density of the internal, sensory, and active states, and attempt to suppress their surprisal or, on average, their entropy. We can use a series of mathematical `inflationary devices' (adding and subtracting the same quantity,so the total is 0, in order to introduce it to the equation) to express this surprisal in terms of its interaction with the external states beyond the blanket.
\begin{align*}
    -\ln p^*(i,s,a) &= \E_{p^*(e | i,s,a)}\big[ -\ln p^*(i,s,a) \big] \\
    &= \E_{p^*(e | i,s,a)}\big[ \ln p^*(e | i,s,a) -\ln p^*(e,i,s,a) \big] \\
    &= \E_{p^*(e | i,s,a)}\big[ \ln p^*(e | i,s,a) -\ln p^*(i,s,a | e) - \ln p^*(e) \big] \\
    &=\underbrace{\E_{p^*(e | i,s,a)}\big[ -\ln p^*(i,s,a | e) \big]}_{\text{Inaccuracy}} + \underbrace{\KL \big[p^*(e | i,s,a) || p(e)\big]}_{\text{Complexity}} \numberthis
\end{align*}
Thus we can see that the flow of autonomous states acts to minimize the inaccuracy (maximize accuracy) and minimize the complexity of the external states with respect to the `particular states' of the system in question. The particular states consist of the sensory, active, and internal states -- i.e. everything except the external states. Parsed into more intuitive terms, we can thus see that the flow of `system' states ($i,s,a$) aim to maximize the `\emph{likelihood}' of the internal states given the external states -- i.e. perform maximum likelihood inference on themselves (c.f. `self evidencing' \citep{hohwy2016self}) -- while simultaneously minimizing the complexity -- or the divergence between the external states given the internal states, and the `prior' distribution over the external states. In short, by re-expressing the flow in information-theoretic terms, we can obtain a decomposition of the entropy term into intuitive and interpretable sub-components which can help us reason about the kinds of behaviours these systems must exhibit. 

Further discussion of the nature and necessity of the Markov Blanket conditions, and the various additional constraints on solenoidal coupling are discussed in detail in the appendix (section $10.3.2$)

\section{Variational Inference}

Variational inference is a method for approximating intractable integrals in Bayesian statistics \citep{feynman1998statistical,jordan1998introduction,ghahramani2001propagation,jordan1999introduction,fox2012tutorial,neal1998view}. Typically, a direct application of Bayes-rule to compute posteriors in complicated systems fails due to the intractability of the log model evidence, which appears in the denominator of Bayes' rule. While there exist numerical or sampling-based methods to precisely compute this integral, they typically scale poorly with the dimension of the problem -- a phenomenon which is known as the curse of dimensionality \citep{goodfellow2016deep}. Variational techniques originated from methods in statistical physics in the 1970s and 1980s \citep{feynman1998statistical}, and were then taken up in mainstream statistics and machine learning in the 1990s \citep{ghahramani2001propagation,beal2003variational,jordan1998introduction} where they have become an influential, often dominant approach for approximating posteriors and fitting complex high-dimensional Bayesian models to data \citep{feynman1998statistical,jordan1999introduction,ghahramani2000graphical,beal2003variational,blei2017variational,kingma_auto-encoding_2013,dayan1995helmholtz}.

The core idea of variational inference is to approximate an intractable inference problem with a tractable optimization problem. Thus, instead of directly computing a posterior distribution $p(H | D)$ where $H$ is some set of hypotheses and $D$ is the data, we instead postulate an approximate or variational distribution $q(H | D; \theta)$ which is often, although not always, parametrized with some fixed number of parameters $\theta$. We then seek to optimize the parameters $\theta$ to minimize the divergence between the approximate and true posterior,
\begin{align*}
\theta^* = \underset{\theta}{argmin} \, \KL[q(H | D; \theta) || p(H | D)] \numberthis
\end{align*}
Unfortunately, this optimization problem is itself intractable since it contains the intractable posterior as an element. Instead, we minimize a tractable bound on this quantity called the variational free energy (VFE) $\mathcal{F}(D,\theta)$,
\begin{align*}
\mathcal{F}(D, \theta) &= \E_{q(H|D)}[\ln q(H | D; \theta) - \ln p(H, D)] \\
&= \KL[q(H | D; \theta) || p(H | D)] - \ln p(D) \\
&\geq \KL[q(H | D; \theta) || p(H | D)] \numberthis
\end{align*}

Since the VFE is simply a divergence between the variational distribution and the generative model $p(D,H)$, it is tractable as we assume we know the generative model that gave rise to the data. By minimizing the VFE, therefore, we reduce the divergence between the true and approximate posteriors, and thus improve our estimate of the posterior.

Secondly, the variational free energy is simultaneously a bound upon the log model evidence $\ln p(D)$, which is precisely the quantity which was intractable to compute due to the implicit integration over all possible hypotheses (or parameters) $p(D) = \int dH p(D | H)p(H)$. However, if we perfectly succeed in matching the variational and true posteriors, then the free energy simply converges to precisely the log model evidence. It is then possible to use this estimate for model comparison and selection \citep{friston2018bayesian,geweke2007bayesian} since it provides a metric to score the fit of a model on the data.
\begin{align*}
\ln p(D) &= \E_{q(H|D; \theta)}[ \ln q(H | D; \theta) - \ln p(H, D)] - \mathcal{F}(D,\theta) \\
&\geq -\mathcal{F}(D,\theta) \numberthis
\end{align*}
The second line follows due to the non-negativity of the KL divergence. The VFE is the foundation of the Free-energy principle as, we shall show, we can interpret self-organizing systems which maintain themselves at a non-equilibrium-steady state to be implicitly minimizing the VFE, and thus performing variational Bayesian inference.

We can gain some intuition for the effects of minimizing the VFE by decomposing it into various constituent terms. Here we showcase two different decompositions which each give light to certain facets of the objective function,
\begin{align*}
    \mathcal{F}(D, \theta) &= \E_{q(H|D ; \theta)}[q(H | D; \theta) || p(H, D)] \\
    &= \underbrace{\E_{q(H | D; \theta)}[\ln p(H,D)]}_{\text{Energy}} - \underbrace{\mathbf{H}[q(H | D; \theta)]}_{\text{Entropy}} \numberthis \\ 
    &= \underbrace{\E_{q(H | D;\theta)}[\ln p(D | H)]}_{\text{Accuracy}} + \underbrace{\KL[q(H | D;\theta)||p(H)]}_{\text{Complexity}} \numberthis
\end{align*}
Here we see that we can decompose the variational free energy into two separate decompositions, each consisting of two terms. The first decomposition splits the VFE into an `energy' term, which effectively scores the likelihood of the generative model under the variational distribution, while the entropy term encourages the variational distribution to become maximally entropic. Essentially, this decomposition can be interpreted as requiring that the variational distribution maximize the joint probability of the generative model (energy), while simultaneously remaining as uncertain as possible (entropy) \footnote{Interestingly, this energy-entropy decomposition is precisely why this information-theoretic quantity is named the variational \emph{free energy}. The thermodynamic free energy, a central quantity in statistical physics, has an identical decomposition into the energy and the entropy.}. The second decomposition -- into an `accuracy' and a `complexity' term -- speaks more to the role of the VFE in inference. Here the accuracy term can be interpreted as driving the variational density to produce a maximum likelihood fit of the data, by maximizing their likelihood under the variational density. The complexity term can be seen as a regularizer, which tries to keep the variational distribution close to the prior distribution, and thus restrains variational inference from pure maximum-likelihood fitting.

\section{Intrinsic and Extrinsic information geometries}

Now, we wish to understand the relationship between the internal states and the external states, which are separated by the blanket states. Given the existence of a blanket, the next move is to define a mapping, denoted $\sigma$ between the most likely internal state and the most likely external state, given a specific blanket state.  While such a mapping is not guaranteed to exist in general, it does under certain conditions -- namely if we assume injectivity between the most likely internal and blanket states \citep{parr2020markov}. For linear OU processes, the $\sigma$ mapping always exists and can be analytically derived relatively straightforwardly \citep{aguilera2021particular,da2021bayesian}.

We define the most likely internal and external states given a blanket state as \footnote{A small assumption introduced here is that the argmax for the distribution of external or internal states given a blanket state is unique.},
\begin{align*}
\mathbf{e}(b) &= \underset{e}{argmax} \,  p(e | b) \\
\mathbf{i}(b) &= \underset{i}{argmax} \, p(i | b) \numberthis
\end{align*}

From this, we can define $\sigma$ to be the mapping that fulfils the following equation,
\begin{align*}
\mathbf{e}(b) = \sigma(\mathbf{i}(b)) \numberthis
\end{align*}
Importantly, we can interpret the output of this function -- the most likely external states given the blanket states -- as parametrizing the mean of a distribution over the external states, as a function of the internal states $q(e; \mathbf{e}(b)) = q(e; \sigma(\mathbf{i}(b)))$. This allows us to interpret the flow of the mean of internal states as parametrising a changing distribution over the external states. 

Crucially, we can say that if any given set of internal states parametrizes a distribution over external states, then the space of internal states effectively represents a \emph{space of distributions} over external states, parametrized by internal states. This space of distributions may be, and usually is, curved and non-euclidean in nature. The field of information geometry provides many mathematical tools to allow us to describe and mathematically characterise such spaces correctly \citep{amari1995information,caticha2015basics}. A key result in information geometry is that the space of parameters of families of exponential distributions is a non-euclidean space with the Fisher Information as its metric. A metric is simply a notion of distance for a given space. For instance, in Euclidean space, the metric is $\sqrt{\sum_i^N x_i^2}$ where $N$ is the dimensionality of the space and the $x_i$s are the coordinate vectors of the space. We can represent general coordinate transformers on spaces with any metric through the use of a metric tensor $\mathbf{G}$.  Essentially, we measure differences between distributions in terms of the KL divergence, and thus if we want to see how an infinitesimal change in the parameters of a distribution results in changes to the distribution itself, we can measure an infinitesimal change in their KL divergence as a function of the infinitesimal change in the parameters. i.e. 
\begin{align*}
\frac{\partial p(x; \theta)}{\partial \theta} = \lim_{\delta \theta \to 0} \KL[p(x; \theta) || p(x; \theta + \delta \theta)] \numberthis
\end{align*}
In the case of the space of parameters of exponential distributions, the metric tensor is the Fisher information, which arises as from the Taylor expansion of the infinitesimal KL divergence between the two distributions. We define $\theta' = \theta + \delta \theta$. Specifically, since there is only an infintesimal change, we can Taylor-expand around $\theta' = \theta$ to obtain,
\begin{align*}
    \KL[p(x;\theta)||p(x;\theta')] &\approx \underbrace{\KL[p(x;\theta)||p(x;\theta)]}_{=0} + \underbrace{\frac{\partial \KL[p(x;\theta)||p(x;\theta')]}{\partial \theta}|_{\theta = \theta'}(\theta - \theta')}_{=0} + \frac{\partial^2 \KL[p(x;\theta)||p(x;\theta')]}{\partial \theta^2}|_{\theta = \theta'}(\theta - \theta')^2 \numberthis
\end{align*}`
Where the first two terms vanish, so we need only handle the second term,
\begin{align*}
   \KL[p(x;\theta)||p(x;\theta')] &\approx \frac{\partial^2 \KL[p(x;\theta)||p(x;\theta')]}{\partial \theta^2}|_{\theta = \theta'}(\theta - \theta')^2 \\
   &= \int p(x; \theta) \frac{\partial \ln p(x;\theta)}{\partial \theta}\frac{\partial \ln p(x;\theta)}{\partial \theta}d\theta \\ 
   &= \mathcal{I} \numberthis
\end{align*}
where $\mathcal{I}$ is the Fisher information. Since the internal states can be interpreted as parametrizing distributions over external states, as parameters, they lie on an information-geometric manifold with a Fisher information metric. This is the extrinsic information geometry. Simultaneously, the internal states also parametrize (implicitly) a second (empirical) distribution over the internal states. This parametrization gives rise to a second information geometry -- the intrinsic geometry, since it represents the relationship the internal states have to the distribution over themselves. Specifically, suppose $\mathbf{i}$ define the sufficient statistics of a density over internal states $p(i; \mathbf{i})$, and $\mathbf{e} = \sigma(\mathbf{i})$ define the sufficient statistics of the variational density over external states $q(e;\mathbf{e})$, then we can see that the internal states in fact parametrize \emph{two} densities and thus partake in two simultaneous information geometries. First, there is a metric defined over the space of internal densities,
\begin{align*}
    \mathcal{I}(\mathbf{i}) = \frac{\partial ^2 \KL[p(i;\mathbf{i})||p(i;\mathbf{i} + \delta \mathbf{i})]}{\partial \mathbf{i}^2}|_{\mathbf{i} + \delta \mathbf{i} = \mathbf{i}}\numberthis
\end{align*}
which is called the \emph{intrinsic} information geometry. And secondly, a metric defined over the space of external densities, parametrized by internal states,
\begin{align*}
    \mathcal{I}(\mathbf{e}) =\frac{\partial^2 \KL[q(e; \mathbf{e})||q(e;\mathbf{e} + \delta \mathbf{e})]}{\partial \mathbf{e}^2}|_{\mathbf{e} + \delta \mathbf{e} = \mathbf{e}} \numberthis
\end{align*}
which is called the \emph{extrinsic} information geometry. These well-defined intrinsic and extrinsic information geometries, allow us to interpret the motion of the internal as also representing motion on the intrinsic and extrinsic statistical manifolds. Crucially, enabling us to make mathematically precise the link between two conceptually distinct ideas -- dynamical motion in space, and variational inference on parameters of distributions. Using this underlying information-geometric framework, in the next section we shall go on to see how we can interpret the dynamics of a non-equilibrium system at NESS as performing approximate variational Bayesian inference on its external environment.

\section{Self-Organization and Variational Inference}
Here we present the key results of the Free-energy principle via the Free-Energy lemma. Specifically, this says, firstly, that the dynamics of the autonomous states can be interpreted as minimizing a free energy functional over the external states, and thus can be construed as performing a kind of elemental Bayesian (variational) inference. This section relies on a fair number of assumptions which are controversial within the FEP community. Here we present the `ideal narrative' where these assumptions are taken as fact. For a more critical discussion of the steps and assumptions in this section, please see the appendix.

We will first consider the general case in terms of the `particular' free energy, which stipulatively assumes that the system obtains the correct posterior at every time-point, rendering the traditional variational bound superfluous, and thus demonstrating that in a way self-organizing systems maintaining themselves at NESS can be construed as performing \emph{exact} Bayesian inference on the generative model they embody through their NESS density. We thus reach a first draft of the key statement of the FEP -- that the dynamics self-organizing systems that maintain themselves at NESS can be interpreted as performing exact Bayesian inference on the external states beyond their blanket or, alternatively, they can be interpreted as approximating approximate (variational) Bayesian inference. 

We then introduce the general case of the \emph{variational free energy}, which is in general a bound upon the marginal NESS density, and we show in the special case of assuming that the variational distribution over external states which is parametrized by the internal states can be approximated by the Laplace approximation, that we can interpret the flow of autonomous states as directly performing a descent upon the variational free energy and thus directly performing variational Bayesian inference. Since we, as the modeller, can specify the variational distribution in any desired way, then this means that this interpretation is potentially tenable for a wide range of systems. The Laplace approximation approximates the variational distribution as a Gaussian where the variance is a function of the curvature at the mean. Intuitively, this assumption is that the Gaussian is tightly peaked around the mean value. In linear systems, this approximation is theoretically well-justified, due to the underlying Gaussianity of the stochastic noise in the system, and the likely concentration of the probability mass around the mean, since the noie distribution is unimodal. However, in nonlinear systems complex multimodal distributions may emerge, even with purely Gaussian noise due to intrinsic nonlinearities of the dynamics, and so this approximation may perform poorly. On the other hand, the Gaussian distribution arises regularly in nature whenever averages over large numbers of independent events are taken c.f. the Central Limit Theorem (CLT), and can thus be considered a natural modelling choice for distribution of the mode of the external states given the blanket, which likely is composed of contributions from a large number of specific external states. 

To recall, we can write the flow of autonomous states $\alpha = (i,a)$ in terms of a gradient descent on the log NESS density of the particular states $\ln p(s,i,a)$ with both dissipative and solenoidal components via the Helmholtz decomposition.
\begin{align*}
f_\alpha = (\Gamma - Q)\nabla_\alpha \ln p^*(s, \alpha) \numberthis
\end{align*}

Then we can define the particular free energy as the variational free energy, where the variational distribution over external states, is stipulatively defined to be equal to the `true' posterior distribution over external states given the particular states $q(e | s,i,a) = p^*(e | s,i,a)$ \footnote{We implicitly assume here that the variational distribution can be stipulated to be of the same family of the true posterior, so that they can match one another}. With this assumption, we can define the particular free energy using the standard form for the variational free-energy
\begin{align*}
\mathcal{F}_{particular} &= \KL[q(e | i,s,a) || p^*(e,i,s,a)] \\
&= \underbrace{\mathbf{E}_{q(e | i,s,a)}[\ln p^*(i,s,a | e)]}_{\text{Accuracy}} + \underbrace{\KL[q(e | i,s,a) || p^*(e)]}_{\text{Complexity}} \\
&= \underbrace{\ln p^*(i,s,a)}_{\text{Evidence}} + \underbrace{\KL[q(e | i,s,a) || p^*(e  | i,s,a)]}_{\text{Bound = 0}} \\
&= \ln p^*(i,s,a) \numberthis
\end{align*}
where the last line follows because the bound is always 0 since we have defined the variational and true posteriors to be the same. Importantly, we see that the particular free energy is then equal to the log of the NESS density over the sensory, internal, and active states. As such, we can rewrite the dynamics of the autonomous states directly in terms of the particular free energy,
\begin{align*}
\frac{d\alpha}{dt} = (\Gamma - Q) \nabla_\alpha \mathcal{F}_{particular}(s, \alpha) \numberthis
\end{align*}

While this may seem like just a mathematical sleight of hand, it demonstrates how systems which maintain the statistical structure of a Markov Blanket at equilibrium can in fact be interpreted as performing variational Bayesian inference with a correct posterior distribution. If, conversely, we relax this assumption somewhat, so that, as is typical for variational inference when the class of distributions represented under the variational density does not include the true posterior, then we retain an approximate relationship. That is, when $q(e | i,s,a;\theta) \approx p(e | i,s,a)$, we obtain,
\begin{align*}
\mathcal{F} &= \KL[q(e |i,s,a) || p^*(e,i,s,a)] \\
&= \ln p^*(i,s,a) + \KL[q(e | is,a) || p^*(e  | i,s,a)] \\
&\approx \ln p^*(i,s,a) \\
&\implies \frac{d\alpha}{dt} \approx (\Gamma - Q) \nabla_\alpha \mathcal{F}(s, \alpha)
\end{align*}

So we can see that in this case, we can interpret the dynamics of the autonomous states as \emph{approximating} approximate Bayesian inference. This is perhaps the most general statement of the FEP -- that the dynamics of a system which maintains the statistical structure of a Markov Blanket at NESS against external dissipative perturbations, can be interpreted as performing approximate variational Bayesian inference to optimize a distribution over the external states of the environment, parametrized by its own internal states. The distinction between variational and particular free energy, with the particular free energy always using the stipulatively correct posterior, while being somewhat a mathematical trick, is also a useful philosophical distinction to draw. In effect, we can think of the system as always performing correct Bayesian inference, simply because the inference is over the system itself, where the generative model of the system is simply its NESS density. Conversely, we can see the approximation arising from the approximate variational distribution as being related to the imperfection of our own understanding of the system as an exogenous modeller. The system is perfectly happy using its Bayes-optimal posterior at all times. A variational distribution distinct from this posterior must be, in some sense, the creature and creation of the modeller, not of the system, and as such the approximations to the dynamics that arise from this approximation is due to the approximations implicit in modelling rather than in the dynamics of the system per-se. It is also important to note that while we have used an approximation sign, in reality the variational free energy is an \emph{upper bound} upon the log model evidence or the particular free energy -- i.e. $\mathcal{F} \geq \mathcal{F}_{particular}$ and the approximate dynamics can be interpreted as driving the system towards the minimization of this bound, and thus increasing the accuracy of the approximation in a manner analogous to the similar process inherent in variational inference.

While in the general case above, the relationship between the dynamics of the system and variational inference is only approximate, if we are only interested in the distribution over the \emph{mode} of the external states -- i.e. the most likely external state configuration -- instead of the full distribution, then the approximation becomes exact and we can directly see that the dynamics of the system do perform variational inference upon the mode of the external states. Here we can see that, in a sense, the maximum-a-posteriori (MAP) modes for the internal states precisely track the MAP modes for the external states and thus, under the Laplace approximation, can be seen as directly performing a minimization of the variational free energy.

Firstly, recall from previously that we had defined the smooth mapping between the modes of the external and internal states given the blanket state, $\mathbf{e}(b) = \sigma(\mathbf{i}(b))$. By applying the chain rule to this function, it is straightforward to derive the dynamics of the external mode with respect to the internal mode,
\begin{align*}
\frac{d \mathbf{e}(b)}{dt} = \frac{\partial \sigma(\mathbf{i}(b))}{\partial \mathbf{i}(b)} \frac{d \mathbf{i}(b)}{dt} \numberthis
\end{align*}
Then, assuming that the mapping is invertible (requiring that the internal states and external states have the same dimensionality), or rather in the general case that it has a Moore-Penrose pseudoinverse, we can express the dynamics of the internal mode in terms of the dynamics of the external mode,
\begin{align*}
\label{change_internal_mode}
\frac{d \mathbf{i}(b)}{dt} = \frac{\partial \sigma(\mathbf{i}(b))}{\partial \mathbf{i}(b)}^{-1} \frac{d \mathbf{e}(b)}{dt} \numberthis
\end{align*}
Similarly, we can derive the expression NESS density over the external mode in terms of the mode of the internal states, which provides a precise mapping, called the \emph{synchronization manifold}, between the two densities, even though they are in fact separated by the Markov Blanket,
\begin{align*}
\label{change_external_by_internal}
\frac{\partial \ln p(\mathbf{e}(b) | b)}{\partial i} = \frac{\partial \ln p(\mathbf{e}(b) | b)}{\partial \mathbf{e}(b)}\frac{\partial \sigma(\mathbf{i}(b))}{\partial i} \numberthis
\end{align*}
Combining Equations \ref{change_external_by_internal} and \ref{change_internal_mode} and using the fact that the dynamics of the external mode, by the marginal flow lemma are, $\frac{d \mathbf{e}(b)}{dt} = (\Gamma_e - Q_e) \nabla_e \ln p(\mathbf{e}(b) | b)$, we can express the dynamics of the internal mode in terms of the marginal NESS density over the external states, thus understanding how the internal states probabilistically track changes in their environment,
\begin{align*}
\label{approximate_bayes_derivation}
\frac{d \mathbf{i}(b)}{dt} &= \frac{\partial \sigma(\mathbf{i}(b))}{\partial \mathbf{i}(b)}^{-1} \frac{d \mathbf{e}(b)}{dt} \\
&= \frac{\partial \sigma(\mathbf{i}(b))}{\partial \mathbf{i}(b)}^{-1} (\Gamma_e - Q_e) \nabla_e \ln p(\mathbf{e}(b) | b) \\
&= \frac{\partial \sigma(\mathbf{i}(b))}{\partial \mathbf{i}(b)}^{-1} (\Gamma_e - Q_e) \frac{\partial \sigma(\mathbf{i}(b))}{\partial \mathbf{i}(b)}^{-1} \frac{\partial \sigma(\mathbf{i}(b))}{\partial \mathbf{i}(b)} \nabla_e \ln p(\mathbf{e}(b) | b) \\
 &= (\Gamma_\sigma  - Q_\sigma) \nabla_i \ln p(\sigma(\mathbf{i}(b)))  \numberthis
\end{align*}
where $(\Gamma_\sigma - Q_\sigma) = \frac{\partial \sigma(\mathbf{i}(b))}{\partial \mathbf{i}(b)}^{-1} (\Gamma_e - Q_e) \frac{\partial \sigma(\mathbf{i}(b))}{\partial \mathbf{i}(b)}^{-1}$. Crucially, this expression allows us to express the dynamics of the internal mode as a gradient descent on the NESS density of the external mode given the blanket, with respect to the mode of the internal states. Fascinatingly, this relationship takes the same general form of the Helmholtz decomposition with separate dissipative $\Gamma_\sigma$ and solenoidal $Q_\sigma$ components which are simply the original dissipative and solenoidal components with respect to the internal states modulated by the inverse of the mapping function $\sigma$. In effect, this implements a coordinate transform between the coordinates of the dynamics of the external states to the coordinates of the dynamics of the mode of the external states, as a function of internal states.

Now we demonstrate how we can interpret this gradient descent on the NESS density of the mode over external states in terms of a direct descent on the variational free energy, and thus as directly and exactly performing variational inference. First, we must define our variational distribution $q(\mathbf{e} | b; \mathbf{i})$ which is a distribution over the modes of external states, given the blanket states, parametrized by the mode of the internal states. Since we are only interested now in distributions over the \emph{mode} of the external states, a reasonable assumption is that it is approximately Gaussian distributed due to the central limit theorem. This means that a Laplace approximation, which is a Gaussian approximation where the covariance is simply a function of the mean, derived via a second order Taylor-expansion of the density at the mode, is a good approximation to use here. We thus define the variational density as,
\begin{align*}
& q(\mathbf{e} | b; \mathbf{i}) = \mathcal{N}(\mathbf{e}; \mathbf{i}, \Sigma(\mathbf{i})) \\
& \text{where} \, \, \Sigma(\mathbf{i}) = \frac{\partial^2 \sigma(\mathbf{i})}{\partial \mathbf{i}^2}^{-1} \numberthis
\end{align*}
Importantly, if we substitute this definition of $q$ into the variational free energy and drop constants unrelated to the variational parameters $\mathbf{i}$, we obtain,
\begin{align*}
\mathcal{F} &= \ln p(\mathbf{i}, b) + \frac{1}{2}tr(\Sigma(\mathbf{i})) \frac{\partial^2 \sigma(\mathbf{i})}{\partial \sigma^2}^{-1} + \ln | \Sigma(\mathbf{i}) | \\
&\implies \frac{\partial \mathcal{F}}{\partial i} = \frac{\partial \ln p(\mathbf{i}, b)}{\partial i}
\end{align*}
The second line follows since this is the only term where $i$ is directly utilized. Then, from this definition, we can see that the variational free energy is actually precisely the gradient term we see in the expression for the dynamics of the internal state mode, thus allowing us to rewrite it as,
\begin{align*}
\label{internal_descent_free_energy}
\frac{d \mathbf{i}}{dt} = (\Gamma_\sigma  - Q_\sigma) \nabla_i \mathcal{F} \numberthis
\end{align*}
After this thicket of mathematics, we thus see a crucial result for the FEP. That, with a Laplace-encoded variational density, we can see that the mode of the internal states precisely tracks the mode of the external states, and the dynamics that allows it to do so are precisely those of a gradient descent on the variational free energy, thus enabling an exact interpretation of the dynamics of the internal states as performing Bayesian inference on the external states. This proof demonstrates the fundamentally Ashbyan nature of self-organization at non-equilibrium steady state, where systems, in order to maintain their steady state, and thus existence as distinct systems, are necessarily forced to engage in some degree of modelling or tracking external states of the environment, in order to counter their dissipative perturbations. Interestingly, this exact relationship to variational inference only emerges when considering the \emph{modes} of the system, not the full distribution over environmental and internal states as was done previously, where we only obtained an approximation to variational inference. Perhaps this is because, in some sense, the system need not perform inference on full distributions, but only on modes. This perhaps makes more intuitive sense within the cybernetic Ashbyan paradigm where, in general, the system is seen as significantly smaller than the environment, and thus simply cannot be expected to encode a fully accurate model of the entire environment which, in the extreme case, includes the entire rest of the universe. Instead, the system simply models and tracks coarse-grained environmental variables such as the mode.

An important additional note is that the gradient descent also contains solenoidal terms based upon $Q_\sigma$. Since these terms are orthogonal to the gradient of the free-energy, they do not affect the ultimate minimum of the descent, however they may alter its rate of convergence, since solenoidal terms encourage additional exploration of the state-space than a simple gradient descent does. This result also means that strictly speaking, a block diagonal $Q$ matrix, or even a state independent $Q$ matrix is not necessary for this derivation as relaxing these assumptions will simply result in additional solenoidal coupling terms in equation \ref{internal_descent_free_energy} but will not change the ultimate minimum of the descent so long as the solenoidal coupling terms remain orthogonal to the gradient of the free energy.

\section{The Expected Free Energy and Active Inference}

\begin{figure}
    \centering
    \includegraphics[scale=1]{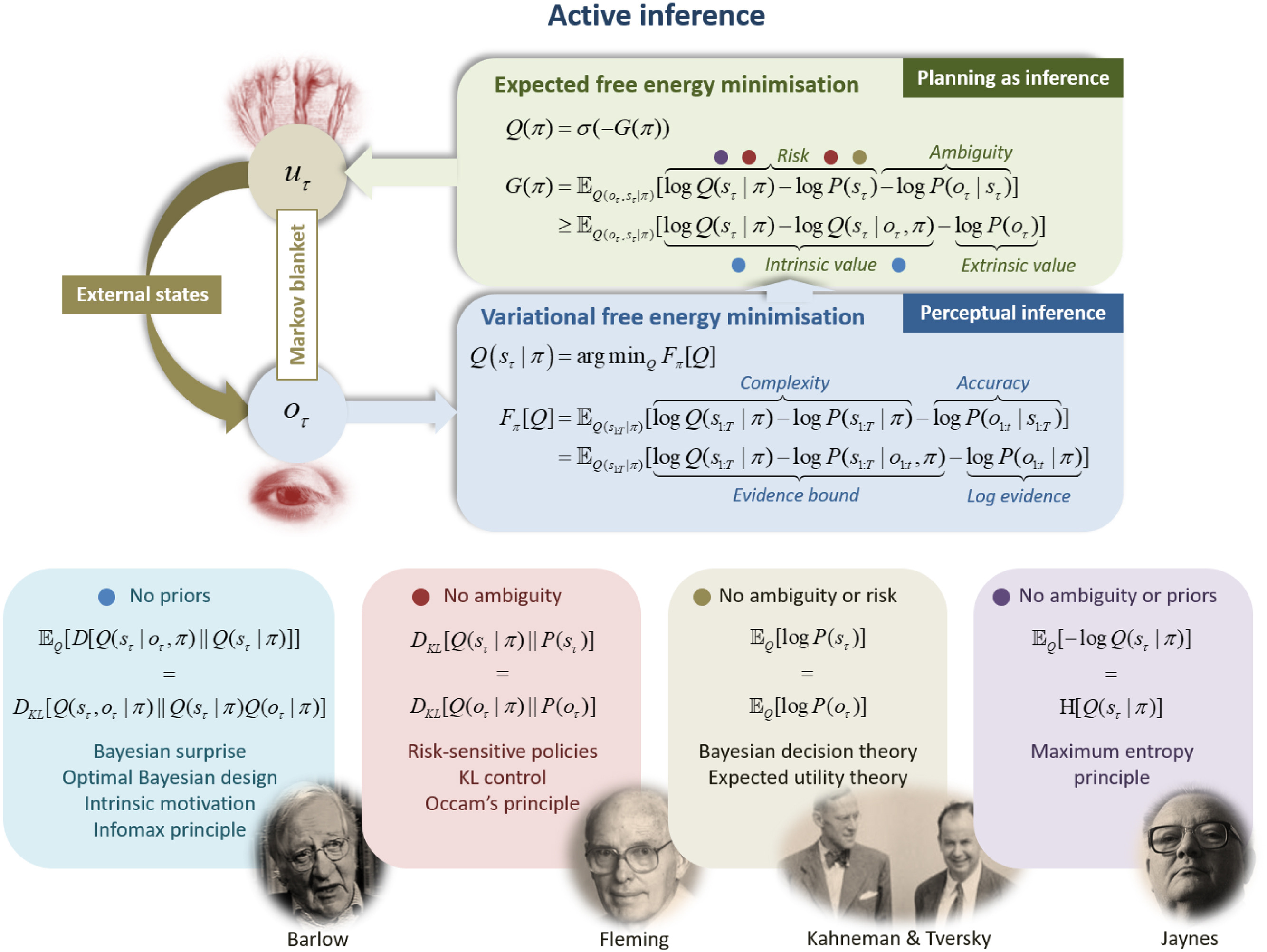}
    \caption{Active Inference and Free Energy Minimization. Top: We see that discrete state space active inference requires two separate minimizations -- one of variational free energy for perception, and one of expected free energy for action selection. Bottom: We see that the EFE functional can be decomposed in various ways to yield a variety of influential objectives which have previously been proposed in the literature. Figure originally in \citet{da2020active}}
    \label{fig:my_label}
\end{figure}
So far, we have only considered the relationship between internal and external states, and observed that the dynamics of the internal state can be considered to be performing a variational gradient descent on the parameters of the variational density over external states. The internal state dynamics exactly follow a variational gradient descent if we assume that the internal states parametrize a Laplacian approximate posterior, or they approximately follow a variational gradient descent if we assume a broader class of variational posteriors. From this, we can interpret the dynamics of the internal states as performing some kind of `perceptual' inference about the causes of fluctuations in the blanket states -- namely, the external states. But what about the active states? How do they fit into this picture? 

First, we recall from the approximate Bayesian inference lemma that we can express the dynamics of the autonomous states (active and internal) in terms of an approximate gradient descent on the variational free energy (Equation \ref{internal_descent_free_energy}). By the marginal flow lemma, if we ignore solenoidal coupling between internal and active states, we can partition this descent into separate (marginal) descents on the internal and the active states, allowing us to write the dynamics of the active states as
\begin{align*}
     \frac{da}{dt} \approx (\Gamma_{aa} - Q_{aa}) \nabla_a \mathcal{F}(s, \alpha) \numberthis
\end{align*}
where $\Gamma_{aa}$ and $Q_{aa}$ are the block matrices corresponding solely to the interactions between active states in the larger $\Gamma$ and $Q$ matrices. Crucially, if we recall the definition of the variational free energy,
\begin{align*}
    \mathcal{F}(i,s,a) &= \E_{q(e;\mu)}[\ln q(e; \mathbf{\mu}) - \ln p^*(e, i,s,a)] \\
    &= \underbrace{\E_{q(e | \mathbf{\mu})}[ -\ln p^*(i,s,a | e)]}_{\text{Inaccuracy}} + \underbrace{\KL[q(e ; \mathbf{\mu})||p^*(e)]}_{\text{Complexity}} \numberthis
\end{align*}
Crucially, the only term in this decomposition that depends on the active states $a$ is the first \emph{inaccuracy} term. Thus, we can straightforwardly write down the dynamics of the active states as,
\begin{align}
    \frac{da}{dt} \approx (\Gamma_{aa} - Q_{aa}) \nabla_a \E_{q(e | \mathbf{\mu})}[ -\ln p^*(i,s,a | e)] \numberthis
\end{align}
Where we can intuitively see that the dynamics of the active states effectively \emph{minimize} inaccuracy (or maximize accuracy). In effect, we can interpret the dynamics of the active states at the NESS density to try to ensure that the variational 'beliefs' encoded by the blanket and internal states of the system are as accurate as possible. Since active states can only influence external states and \emph{not} internal states, the way this is achieved is by acting upon the external states to bring them into alignment with the beliefs represented by the internal states -- hence \emph{active inference}.

While this provides a good characterisation of the dynamics of the system at equilibrium, we are often also interested in the properties of dynamical systems as the self-organize \emph{towards} equilibrium. Specifically, we wish to characterise the nature of the active states during this process of self-organization, so that we can understand the necessary kinds of active behaviour any self-organizing system must evince. To begin to understand the nature of this self-organization we first define another information theoretic quantity, the \emph{Expected Free Energy} (EFE) which serves as an upper-bound on surprisal throughout the entire process of self-organization, with equality only at the equilibrium itself. Since we have this upper-bound, we can interpret self-organizing systems away from equilibrium, by following their surprisal dynamics as approximating expected free energy minimization, using logic directly analogous to the approximate Bayesian inference lemma. Conversely, turning this logic around lets us \emph{construct} self-organizing systems by defining some desired NESS density, and then prescribing dynamics which simply minimize the EFE.

To handle systems away from equilibrium, we define some new terminology. We define $p(e_t, i_t, s_t, a_t | e_0, i_0, s_0, a_0)$ to be the probability density over the variables of the system at some time $t$, which depends on some set of initial conditions $e_0, i_0, s_0, a_0$. To simplify, we average over the external initial condition and only represent the particular initial condition $\pi_0 = (i_0, s_0, a_0)$. Next we define the expected free energy $\mathcal{G}(\pi)$ similarly to the variational free energy, but with the current-time predictive density taking the place of the approximate variational posterior, and the NESS density taking the place of the generative model.
\begin{align*}
    \mathcal{G}(\pi) &= \E_{p(e_t, \pi_t) | \pi_t)}[\ln p(e_t | \pi_t, \pi_0) - \ln p^*(e, \pi)] \\
    &= \underbrace{\E_{p(e_t, \pi_t) | \pi_t)}[-\ln p^*(\pi | e)]}_{\text{Ambiguity}}+ \underbrace{\KL[p(e_t | \pi_t, \pi_0)||p^*(e)]}_{\text{Risk}} \numberthis
\end{align*}
We see that the EFE mandates the minimization of both ambiguity (i.e. avoiding situations which are heavily uncertain) and risk (avoiding large divergences between the current state density and the equilibrium state. It is straightforward to see that the EFE is an upper bound on the expected predictive surprisal at any time-point, by using the fact that the KL-divergence is always greater than or equal to 0,
\begin{align*}
    &\KL[p(e_t, \pi_t | \pi_0)||p^*(e, \pi)] \geq 0 \\
    &\implies \mathcal{G}(\pi_t) + \E_{p(e_t, \pi_t) | \pi_t)}\ln p(\pi_t | \pi_0)] \geq 0 \\
    &\implies \mathcal{G}(\pi_t) \geq - \E_{p(e_t, \pi_t) | \pi_t)}[\ln p(\pi_t | \pi_0)]
\end{align*}
Similarly, it is straightforward to see that at equilibrium the EFE simply becomes the surprisal.
\begin{align*}
    \KL[p(e_t, \pi_t | \pi_0)||p^*(e, \pi)] &=   \mathcal{G}(\pi_t) + \E_{p(e_t, \pi_t) | \pi_t)}[\ln p(\pi_t | \pi_0)]  =0 \\
    &\implies \mathcal{G}(\pi_t) = - \E_{p(e_t, \pi_t) | \pi_t)}[\ln p(\pi_t | \pi_0)] \numberthis
\end{align*}
Since this is the case, we can understand the EFE as effectively quantifying the discrepancy between the current predictive density and the equilibrium. Because of this, we can see that the EFE is necessarily a Lyapunov function of self-organizing dynamics, and it makes sense to interpret self-organizing dynamics under a Markov blanket as minimizing the EFE. Conversely, if one wants to define a set of dynamics that self-organize to some given attractor $p^*(e,i,s,a)$ then one simply needs to define dynamics that minimize the EFE to achieve convergence to the equilibrium (which may be a local minimum). 

Taking this converse approach allows us to move from simply providing an interpretative characterisation of given dynamics in terms of inference, and move instead to constructing or defining systems, or agents, which can achieve specific goals. This approach is taken in the literature on active inference process theories \citep{friston2012active,friston2015active,friston2017active,da2020active} where instead of simply describing a given stochastic differential equation, we instead consider the NESS density to be the \emph{preferences} or \emph{desires} of the agent often represented as a Boltzmann distribution over environmental rewards $p^*(e,i,s,a) = exp(-r(e))$ and the active states (the agent's actions) being computed through a minimization of the EFE, with this minimization either taking place directly as a gradient descent in continuous time and space \citep{friston2009reinforcement} or else as an explicit model-based planning algorithm as in the discrete-time and discrete-space formulation \citep{friston2017process,tschantz2020reinforcement,millidge_deep_2019,millidge2020relationship}.





\section{Philosophical Status of the FEP}
It is worth stepping back from the mathematical morass at this point to try to define at a high level what kind of theory, philosophically speaking, the FEP is, and what kind of claims about the world it makes. There have been numerous debates in the literature about whether the FEP is `falsifiable', or whether it is `correct', and whether or not it makes any specific, empirical claims \citep{williams2020brain,andrews2020math}. However often debates on this matter are obscured or confused by the challenging and deep mathematical background required for a full understanding of the specifics of the FEP. It is clear from the mathematics that the main strand of the FEP offers only an `interpretation' of already extant dynamics. In short, FEP presupposes the existence of the kinds of dynamics it wishes to make sense of -- dynamical systems which organize themselves into a non-equilibrium steady state, and which maintain the requisite statistical independency structure of the Markov Blanket condition. Once these conditions are satisfied, the FEP gives an interpretation of the dynamical evolution of such a system as performing an kind of variational Bayesian inference whereby the internal states of the system (defined by the Markov Blanket partition) can be seen as inferring or representing external states which are otherwise statistically isolated behind the Markov Blanket. Crucially, the FEP, in its most general formulation does not make any specific predictions about the dynamics of the system. It offers an interpretation only. While systems that implement the FEP can be derived, and several process theories have been explicitly derived from within the FEP framework \citep{friston2005theory,friston2015active}, all such theories necessitate making specific and ultimately arbitrary modelling choices, such as of the generative model and variational density. Such choices sit below the level of abstraction that the mathematical theory of the FEP exists at. The FEP, at its core, only offers a mathematical interpretation of certain dynamical structures.

The FEP is often compared and analogised to the principle of least action in physics \citep{lanczos2012variational} which allows one to describe many physical processes (although not all) as minimizing the path integral of a functional called the `action' over a trajectory of motion \citep{sussman2015structure}. This argument is often used to claim, in my opinion correctly, that the FEP is a mathematical `principle' or interpretation and therefore cannot be falsified or empirically tested. In my opinion, however, the principle of least action is, in its philosophical status, not directly analogous to the FEP. While the relationship between the path integral of the action and the dynamics prescribed by the Euler-Lagrange equations is simply a mathematical truth, the principle of least action itself, as applied to physics is contains a fundamentally empirical and falsifiable claim -- that physical systems in the real world can be well described through its own mathematical apparatus -- that is of dynamics derived from minimizing an action. This claim is in principle falsifiable. Not all dynamical systems can be derived from least action principles. If physical systems predominantly came from the class that cannot be so derived, the principle of least action in physics would be effectively falsified, and the mathematical apparatus underlying it would have become nothing more than an arcane mathematical curiosity. So far as we know, there is no a-priori reason why much of physics can be so well understood through action principles, and indeed there are areas of physics -- such as statistical mechanics and thermodynamics, and dissipative non-conservative systems in general -- which \emph{cannot} (so far) be described in these terms.

It appears a closer physics analogy to the FEP might be one direction of Noether's theorem. Noether's theorem proves a direct correspondences between symmetries or invariances in a given system, and conservation laws. For instance, in physical systems, time-translation symmetry implies the conservation of energy, and rotational symmetry (of the underlying euclidean space, not any given object within it) implies the conservation of angular momentum. The FEP, similarly, tries to show a correspondence between the dynamics of a certain kind of system (NESS density, Markov Blanket conditions) and the dynamics of variational Bayesian inference. Interestingly, while the `forward' direction from the NESS density and Markov Blanket conditions handled by the FEP, the reverse direction -- i.e. whether the presence of Bayesian inference dynamics implies any kind of statistical structure upon the dynamics of the system remains unclear, and this is likely a fruitful direction for further theoretical work. Noether's theorem, unlike the principle of least action, matches more closely than the principle of least action since it only specifies correspondences between certain kinds of mathematical objects (symmetries and conservation laws) just as the FEP only specifies a correspondence between dynamical flows at NESS of a system with a Markov Blanket, and the gradient flows on the variational free energy.

While its status as a mathematical principle and interpretation only can shield the FEP from the possibility of an empirical `falsification', this does not mean that the theory is not subject to some kind of implicit intellectual review. Much of the core motivation behind the FEP has been to try to derive universal properties of the kind of biological self-organizing systems which give rise to structured behaviour including relatively `high level' processes such as the perception-action-loop, explicit perception and inference about the causes of the external world and, ultimately, prospective inference and planning. For instance, much of the FEP literature has been focused on and applied to understanding brain function \citep{friston2008hierarchical,friston2015active,friston2017process}. This ambition renders the FEP open to questions about its `applicability', if not its falsifiability. The FEP imposes relatively stringent conditions that dynamical systems must satisfy for the logical steps in the FEP must hold. Although the precise list of assumptions required is not entirely clear, so far in the literature this appears to include at least:
\begin{itemize}
\item that the system in question can be adequately represented as a Langevin equation (i.e. the system is Markov and does not depend on history) with additive white Gaussian noise.
\item that the dynamical system as a whole have a well-defined NESS density (including over the external states)
\item that the system obey the Markov Blanket conditions, which are, in general, relatively restrictive about the kinds of flows that are possible, and appear to have become more restrictive in \citet{friston2020some}, which precludes any solenoidal coupling between active and sensory states
\item that there is an injective mapping between the most-likely internal state given the blanket and the mode of the distribution of external states given the blanket states, which is additionally smooth and differentiable (this is required for the dual-aspect information geometry, and thus the identification with Bayesian inference)
\end{itemize}
These conditions are quite strict about the class of systems that the FEP can apply to, and it is unclear if `real systems' of the kind of the FEP desires to explain -- such as biological self-organization, and especially brains, can fulfil them. If it turns out that such systems flagrantly violate the conditions for the FEP, then the FEP cannot be said to apply to them and thus cannot be of use in understanding them, even as an interpretatory device. In this case, the FEP would fail the applicability criterion, and would cease to be particularly useful for its original goals of neuroscience, even if it remains not technically falsified and does, in fact, apply to some obscure mathematical class of dynamical systems. Importantly, many of the assumptions of the FEP, when interpreted strictly, do not appear to hold in general for complex biological systems such as brains. For instance, to take extreme but illustrative examples, it is clear that no biological system is ever in a true non-equilibrium steady state, since eventually all such organisms will eventually age and die, and indeed eventually the entire universe will likely eventually decay to a thermodynamic equilibrium state. Additionally, the Markov Blanket assumption is directly violated by things such as x-rays (and indeed gravity) which can directly interact with `internal states' of the brain, such as neurons, without first passing through the Markov Blanket of the physical boundaries of the brain and the sensory epithelium. As such, for a real physical system, we must take the assumptions of the FEP to be only approximations, which hold locally, or approximately, but not for all time and with complete perfection. It remains to be seen, and empirically investigated if possible, the extent to which the mathematical interpretations and logical statements of the FEP remain robust to such slight relaxations of its core assumptions.

While the FEP provides a mathematical interpretation of certain kinds of dynamics in terms of inference, it also, largely, remains to be seen whether such an interpretation is useful for spurring new ideas, questions, and developments within the fields the FEP hopes to influence -- such as neuroscience, cognitive science, and dynamical systems theory. Returning to our anaologies of the least action principle and Noether's theorem, while both of these mathematical results only provides interpretations of known dynamics, by operating at a high level of abstraction they provide powerful capabilities for generalization. For instance the principle of least action allows for dynamics to be derived, via the Euler-Lagrange equations, directly from the high level specification of the action. For instance, potentially new or counterfactual laws of physics can be derived simply by postulating a given Lagrangian or Hamiltonian and then working through the mathematical machinery of the principle of least action to derive the ensuing dynamics. Additionally, by investigating equivariances in the action, one can often understand the kinds of invariances and degrees of freedom that exist in the actually realized dynamics. Similarly, Noether's theorem allows one to play with setting up certain conserved quantities or symmetries a-priori, and then work out precisely the consequences that these entail for the dynamics. 

It is currently unclear to what extent the FEP offers such powerful advantages of abstraction and generalization. This is largely due to the FEP being immature as a field compared to the cornerstones of classical physics, and the majority of the research effort so far has gone into making the theory precise rather than deriving consequences and generalizations from it, but there are some promising initial signs which have just begun to emerge in the literature of the power the FEP perspective offers. From a practical perspective, the FEP appears to offer a number of novel techniques. Firstly, given a desired NESS density, the free energy lemma provides a straightforward way of deriving dynamics which will necessarily reach that density, due to the fact that the variational free energy becomes a Lyapunov function of the system as a whole. This approach has strong potential links to Markov-Chain-Monte-Carlo methods in machine learning and statistics, which aim to approximate an intractable posterior distribution by the time evolution of a Markov process \citep{metropolis1953equation,neal2011mcmc,betancourt2017conceptual,chen2014stochastic,brooks2011handbook}. The FEP provides a new perspective on such systems as fundamentally performing variational Bayesian inference, and may in future be used to develop improved algorithms in this domain, akin to the developments of Hamiltonian \citep{betancourt2013generalizing} and Riemannian MCMC \citep{girolami2011riemann} methods. For instance, there is much potential in the idea of solenoidal flow speeding up convergence to the desired equilibrium density \citep{ma2015complete}. Conversely, the FEP, through the Helmholtz decomposition, may additionally provide tools for \emph{inferring the eventual NESS density} given a specific set of dynamics \citep{ma2015complete,friston2019particularphysics}. This would allow for an analytical or empirical characterisation of the ultimate fate of a system, and allow for characterising different kinds of systems purely, such as whether equilibrium or non-equilibrium purely by their dynamics far from equilibrium. 

A second strand of potentially directly useful research which has begun to arise from the FEP is empirical and statistical methodologies for defining, computing, and approximating Markov Blankets. This implies the ability to infer the statistical independency structure of the dynamics either from analytical knowledge of the dynamics or from observed trajectories. There are already two approaches to achieve this in the literature. One which utilizes graph theory in the form of the graph Laplacian to infer nodes of the Markov blanket based on the parents, and children of parents of the largest eigenstates of the Jacobian \citep{friston2013life,palacios2017biological}. A second approach directly uses the Hessian of the dynamics to attempt to read off the conditional independency requirements it implies \citep{friston2020parcels}. These approaches may have substantial merit and utility in understanding the effective statistical independency structure of complex dynamical processes, especially questions regarding functional independence in the brain. This strand of research heavily relates to the question of \emph{abstraction} in dynamical systems -- namely, whether complex systems can or cannot be straightforwardly partitioned into independent `subsystems' which can then be abstracted over. For instance, the ideal would be the ability to, given a complex high-dimensional dynamical system, parse this system into individual `entities' (separated by Markov blankets) which interact with each other according to another set of (hopefully simpler) dynamical rules. This would allow for an automatic procedure to transform a high dimensional complex system into a simpler, low-dimensional approximate system more amenable for analysis and, ultimately understanding.

Finally, it is clear that the process theories inspired by the FEP, although epistemologically they stand apart from the FEP, neither requiring not supporting its validity, have had substantial influence and impact within theoretical neuroscience, where they have been productively applied to understanding a wide range of behavioural and neuronal phenomena.

\section{Discussion}

In this paper, we have aimed to clarify the core logical steps of the FEP, and explain the mathematical apparatus underlying it. At its core, the FEP is a simple theory, which relates the dynamics of a dynamical system with a non-equilibrium steady state, to the dynamics of a variational Bayesian inference process which minimizes a variational free energy functional. The key logic required for this identification is simply the Helmholtz decomposition of the dynamics into a gradient descent on a potential function, and then the Laplace approximation of the variational density which enables the identification of the potential with the variational free energy. However, it is important to realize that the FEP prescribes an \emph{interpretation} of certain kinds of system behaviour only -- it cannot naturally be used in a forward direction to make causal predictions about the evolution of a system, since the dynamics of the system which would be used to make such a prediction must intrinsically be known before the variational free energy functional can be identified. However, the FEP does provide a straightforward mathematical machinery for the design and creation of systems which fulfill its postulates, by providing a mathematical mechanism to translate a desired stationary density with a Markov Blanket conditional independence structure to a set of dynamics which will maintain it. 

While throughout the main body of this paper we have largely endeavoured to present the FEP in the most charitable light, there is still a substantial amount of controversy over the correctness and interpretation of the FEPs mathematical claims -- see \citet{biehl2020technical} and \citet{ aguilera2021particular}. A crucial limitation of the FEP is that many of the conditions required for its derivations are quite restrictive and may significantly limit the kinds of system the FEP can reasonably be applied to. Additionally, many of the claims made in the literature about the power and generality of the FEP are somewhat overstated given these conditions, as well as the FEPs entirely interpretational theoretical status. An important question remains as to the degree to which the assumptions required by the FEP can be slightly relaxed, so as to admit a significantly larger class of potential systems into consideration, without totally destroying the core claims of the FEP as to systems being able to be interpreted as performing variational inference. By presenting the reader with a mathematically detailed, yet ideally intuitive and fairly straightforward presentation of the key concepts of the FEP, we hope to enable them to engage more deeply with the technical literature discussing and debating these questions, as well as to understand on a deep level simply what the FEP does and does not do.
\bibliography{cites}

\section{Appendix}

\subsection{General Review of Assumptions Required for the FEP}

Here we provide a general overview and short discussion of every assumption required at each stage of the FEP. Many of these assumptions have more detailed discussions in subsections of this appendix. Ultimately, the overall picture that emerges is that the FEP requires many assumptions to work, and it is unlikely that all of them can be fulfilled by the kinds of complex self-organizing systems that the FEP ultimately `wants' to be about -- such as biological self organization and, ultimately, brains. However, this does not mean the FEP is useless as many of its assumptions may be `approximately', or `locally' true over small enough time periods. This is not necessarily a bad thing -- almost all of the sciences ultimately use simplified models to try to understand their ultimate objects of study in a more tractable way. The FEP is simply continuing that tradition, but if we do this, we need to make explicit the key distinction between the model and the reality or, more memorably, the map and the territory.

The first set of key assumptions that the FEP makes comes through the definition of the kinds of stochastic dynamical systems that it works with. Specifically, we make the following assumptions about the form of the dynamics we deal with,
\begin{itemize}
    \item The system as a whole can be modelled as a Langevin SDE of the form $\frac{dx}{dt} = f(x) + \omega$
    \item The noise $\omega$ is Gaussian with 0 mean and a covariance matrix $2\Gamma$.
    \item The noise is \emph{additive} to the dynamics
    \item $\Gamma$ does not change with time
    \item $\Gamma$ has no state dependence (no heteroscedastic noise)
    \item $\Gamma$ is a diagonal matrix (each state dimension has independent noise)
    \item The dynamics $f(x)$ do not themselves change with time.
\end{itemize}
We also must make the following assumptions about the system as a whole,
\begin{itemize}
    \item The system is \emph{ergodic}, which means that state and time averages coincide or, alternatively, that there must be some probability of ultimately reaching every part of the system from every other part.
    \item The system possesses a well characterized non-equilibrium-steady state density (NESS), which does not change over time
    \item Once the system reaches this NESS density it cannot escape it -- there is no metastability or multiple competing attractors. This does not mean, however, that the NESS attractor cannot be complex itself and can contain limit cycles.
\end{itemize}
These assumptions setup the basic formalism we wish to consider. From here, we then apply the Ao decomposition to rewrite the dynamics in the form of a gradient descent on the log of the potential function with dissipative and solenoidal components $f(x)= (\Gamma - Q)\nabla_x \ln p^*(x)$. To be able to implement this decomposition requires,
\begin{itemize}
    \item The dynamics function $f$ be smooth and differentiable,
    \item There exists a steady state distribution $p^*(x)$ to serve as the potential function.
\end{itemize}

Now, we apply the Markov Blanket conditions at the NESS density,
\begin{itemize}
    \item The state space $x$ can be partitioned into a set of four states -- internal $i$, external $e$, active $a$ and sensory $s$ which, at the NESS density fulfill the following conditional independence relationships:
    $p^*(x) = p^*(e | s,a)p^*(i | s,a)p^*(s,a)$.
    \item We thus require \emph{all partitions} to be at NESS, including the \emph{external states}. This means that the environment also has to be at steady state, not just the system.
    \item We often assume no solenoidal coupling between internal and sensory states (internal states do not directly act on sensory states -- only the external states do), nor between active and external states (active states drive the external state but are not driven by it). Mathematically this corresponds to $Q_{s,i} = 0$,$Q_{e,a} = 0$. 
\end{itemize}
Given the Markov Blanket conditions hold, we can then begin to move towards the free energy lemma. To begin with, we must first assume,
\begin{itemize}
    \item There is a unique argmax $\boldsymbol{e}, \boldsymbol{i}$ exists for both internal and external states for every blanket state $b$.
    \item That there exists a function $\sigma$ which maps from $\boldsymbol{i}$ to $\boldsymbol{e}$
    \item That $\sigma$ is invertible
    \item That $\sigma$ is differentiable    
    \item For the particular free energy, we assume that the variational posterior $q(e ; \boldsymbol{e})$ is equal to the true posterior, and thus that the true posterior can be represented by one sufficient statistic ($\boldsymbol{e}$).
\end{itemize}
These assumptions on $\sigma$ are quite restrictive. A more detailed discussion of what these assumptions require can be found in Appendix $10.4.1$.

Finally, to reach the free energy lemma, we must make the following assumptions,
\begin{itemize}
    \item The dynamics of the sufficient statistic of external states $\boldsymbol{e}$ follows the same (Ao-decomposition) dynamics as the external states themselves
    \item The variational distribution $q(e;\boldsymbol{e})$ is a Laplace distribution (Gaussian) with a fixed covariance $\Sigma$ as a function of $\boldsymbol{e}$.
    \item The variational covariance $\Sigma$ is not an implicit function of the blanket states
\end{itemize}
This first assumption has come under heavy controversy. These additional assumptions pertain to the Laplace approximation, but the final assumption here appears to go beyond what is typically required by variational Laplace.

\subsection{Assumptions on the Form of the Langevin Dynamics}

The FEP formulation makes reasonably strong assumptions about the nature of the dynamics that it models -- restricting them to the form of stochastic dynamics which can be written as a Langevin equation with additive Gaussian noise. While the assumptions on the dynamics function are not that strong, only requiring differentiability and time-independence, the restrictions on the noise in the system are quite severe.

Firstly, it is important to note that using additive white noise, while a common modelling assumption due to its mathematical simplicity, nevertheless imposes some restrictions on the kind of systems that can be modelled -- especially as complex self organizing systems typically evince some kind of colored smooth noise, as well as often power-law noise distributions which are associated with self-organized criticality \citep{ovchinnikov2016introduction}.

However, the further assumptions on the $\Gamma$ covariance matrix -- that it is diagonal, state-independent, and time-independent -- are also strong additional restrictions. Specifically, this means that the noise to every dimension in the system is completely independent of any other dimension, and that the noise is constant at every point throughout the state space and throughout time.
\subsubsection{Ito vs Ao vs A-type interpretation}

A subtle point raised by Manuel Baltieri (private correspondence) is that the formulation of the FEP in \citet{friston2019particularphysics} is inconsistent between a Stratonovich and an Ito interpretation of the relevant SDEs. This has no impact in the case of non-state-dependent noise, when the two interpretations coincide, but could potentially impact generalizations of the theory to include state dependent noise. 

\subsubsection{Ergodicity and the Ao Decomposition}

The Ao decomposition requires both that the dynamics possess a consistent non-equilibrium steady state density (which forms the potential function) and also that the dynamics are ergodic. Additionally, this ergodicity assumption is implicitly used in the Bayesian mechanics, which allows expectations of the surprisal to be taken and interpreted as entropies, and thus to ultimately derive an interpretation of the dynamics in terms of accuracy and complexity. In general, for many biological and self-organizing systems, ergodicity does not hold and such systems typically exhibit substantial amounts of path dependence and irreversibility. This means that on a strict reading, for most systems the FEP desires to model, the ergodicity assumption does not hold. However, it may still be possible to describe ergodicity as holding 'locally' in the small region of the state space around the NESS density and this may be sufficient for approximate version of the FEP to hold, although the resistance of the FEP to slight perturbations of its assumptions remains unclear.

\subsection{The Markov Blanket Condition}

\subsubsection{Functional vs Statistical Connectivity and the Markov Blanket Condition}

A subtle conceptual issue has been recently raised by \citet{aguilera2021particular} as to the precise meaning of the Markov blanket condition for real systems. Specifically, intuitively the Markov blanket is presented as being a kind of boundary between the internal and external states, and is often presented as a literal boundary -- for instance the cell wall and sensory epithelia of a bacterium, or the sensory epithelia of a brain as opposed to its external environment \citep{friston2019particularphysics,da2021bayesian,parr2020markov}. However, this intuition subtly conflates two types of connectivity -- \emph{functional} connectivity and \emph{statistical} connectivity. Functional connectivity is the literal causal connectivity in the world. For instance, to reach the inside of a bacterial cell, a molecule on the outside must pass through the cell wall. This functional independence in this way can be represented by sparsity (zeroes) in the dynamics matrix $f$. of the whole system. Statistical connectivity, on the other hand, is concerned only with the statistical independencies between variables given the blanket states and is represented by sparsity in the Hessian of the NESS density. Importantly, except under very restrictive conditions, it has been demonstrated by \citet{aguilera2021particular} that these two senses of connectivity are independent of one another -- in that statistical independence does not imply functional separation and vice versa. 

To see why this is the case, imagine the case of water molecules osmosing into the bacterial cell. Here, we assume, there is no functional direct functional connection between inside and outside the cell wall -- i.e. water molecules cannot `teleport' from outside the bacterium directly into it. However, through this process of osmosis, the internal and external states slowly become correlated with each other, as the concentrations of water molecules on both sides of the boundary equalize, thus implying in this case that a functional separation does not necessarily imply a statistical separation. Conversely, imagine taking the densities of two adjacent patches of gas as our variables. If this gas is at thermodynamic equilibrium, then the density in the two patches will be, on average the same, however any tiny fluctuations in density will be purely random and uncorrelated. Thus, even though these adjacent patches have a functional connection (molecules can directly move from one patch to another), they lack a statistical connection.

An important note is that while the FEP generally bases its intuitions about the Markov Blanket in functional terms, the actual mathematical definition is purely statistical. This can lead to confusion about what are the kinds of systems that the FEP actually models and the dynamics they have.

In general functional connectivity expressly precludes statistical independence because, for any noise in the system, as long as there is some causal path between two elements, even if they are separated by a blanket states, noise will propagate through the system and thus internal and external will tend to become correlated with one another. Indeed, this development of correlations over time between internal and external states in the absence of functional connectivity is precisely what we intuitively mean when we think about systems `accumulating knowledge' about their environments, and thus seems important for questions about how systems can learn to model, predict, and infer various facts about the external world. 

However, as originally noted by Martin Biehl in private discussion, the Markov Blanket condition appears to explicitly \emph{preclude} this kind of knowledge accumulation inside the internal states, requiring that all knowledge be held in the blanket states. It may indeed turn out to be the case that defining Markov Blankets in the intuitive sense in terms of functional independencies instead of statistical independencies may lead to a better description of precisely the kinds of knowledge accumulating processes that appear to be important in performing inference.

\subsubsection{The Real Constraints on Solenoidal Coupling?}

While the Markov blanket conditions only explicitly disallow solenoidal coupling directly between the internal and external states -- $Q_{i,e} = 0$, the free energy lemma as stated in Equation \ref{internal_descent_free_energy} appears to require a significantly greater reduction of solenoidal coupling. Specifically, the free energy lemma requires that, for a straightforward identification of the surprisal with the free energy, that the form of the dynamics for each marginal subset of states in the partition take the same form as the dynamics of the full set of states $x$. Specifically, this means that \emph{all} solenoidal coupling between the subsets must be suppressed, since if they were not then, by the marginal flow lemma, there would be additional solenoidal coupling terms in Equation \ref{internal_descent_free_energy}, which would complicate the relation to Free-energy minimization with additional solenoidal terms. As such, for the free energy lemma, as currently presented, we appear to have the extremely strong condition of the diagonality of $Q$, where each subset in the Markov Blanket is only allowed solenoidal interactions with itself.

It is important to note that this restriction is significantly stronger than those required just by the Markov Blanket condition, and indeed is stronger even than the flow constraints proposed in \citet{friston2020some}. While this does not entirely rule out any interactions between different subsets of the Markov blanket, it does mean that all interactions have to be mediated through the gradient term, since both the $\Gamma$ and $Q$ matrices are assumed to be diagonal.

However, it is important to note that if these stringent implicit assumptions on solenoidal coupling in Equation \ref{internal_descent_free_energy} were relaxed, there would be additional solenoidal coupling terms in the equation. However, these terms would be orthogonal to the gradient of the free energy, and thus not materially impact the ultimate minimum of the descent, although they would alter the dynamics of actually reaching the minimum significantly. Specifically, this means that the FEP would instead predict a solenoidal `swirling' kind of gradient descent instead of a direct steepest descent for systems with significant solenoidal couplings.

\subsection{Assumptions of the Free-Energy Lemma}
\subsubsection{The $\sigma$ function}

The existence and general properties of the $\sigma$ function have also drawn much controversy from within the community. Specifically, it is not at all clear that this function exists in the general case, for arbitrary dynamics functions $f$ and conditional NESS distributions $p^*(e | b)$ and $p^*(i | b)$. In later papers it is assumed to exist under the condition of injectivity between $\boldsymbol{e}$ and $\boldsymbol{i}$. In effect, this means that there must be a unique mapping between $\boldsymbol{e}$ and $\boldsymbol{i}$ for all blanket states -- i.e. that for every blanket state, if the argmax of the internal states is $\boldsymbol{i}$, then the argmax of the external states must be $\boldsymbol{e}$. Additionally, there must be a corresponding (and separate) external argmax for every internal argmax. There may, however, be some external argmaxes with no corresponding internal argmaxes (although the converse condition does not hold). This requires that the dimensionality of the external states but greater than or equal to the dimensinoality of the internal states -- which should generally hold for most reasonable systems where we can safely assume that the environment is larger than the system itself. his injectivity condition also guarantees invertibility in the case that the internal and external state spaces are of the same dimension. It is also possible to use the Moore-Penrose pseudoinverse for the case where the external state space is larger, at the cost of the free energy lemma becoming approximate instead of exact.

The differentiability of the $\sigma$ function is a more stringent condition. In many cases this is unlikely to be met, since the argmax functions which the $\sigma$ function maps between are generally nondifferentiable. It remains unclear to what extent differentiable $\sigma$ functions can exist in systems of interest.

However, it can be straightforwardly shown that in the simple case of linear OU processes, that both the $\sigma$ function exists and that it is analytically calculable \citep{aguilera2021particular,da2021bayesian}. Moreover, in the general case, it is possible to obtain an approximate $\sigma$ function by running a regression between the internal and external modes which is often what is done in practice for nonlinear systems \citep{friston2013life}.
\subsubsection{The Dynamics of the Sufficient Statistics e}

An additional important assumption necessary for the free energy lemma, is that the dynamics of the sufficient statistics of the external mode follow the same dynamics as the external states generally -- see Equation \ref{approximate_bayes_derivation}.  This assumption turns out to be crucial to the free energy lemma which relies heavily in the fact that the dynamics of the sufficient statistic $\boldsymbol{e}$ can be written as a gradient descent on the log surprisal -- which can then be expressed in terms of a free energy under the Laplace approximation.

This assumption is also problematic and has been the source of controversy within the community. The extent to which this assumption is justified remains unclear. Specifically, it appears to rule out the use of arbitrary functions $\xi$ (to be discussed in the next section) to parametrize the external sufficient statistic (although not the internal sufficient statistic). The assumption effectively holds to the extent to which one can describe the sufficient statistic as equal to some external state $\boldsymbol{e}(b) \approx e$, which may occur often for the argmax but not necessarily always. It remains to be seen whether the argmax is in fact the optimal such function -- which is dependent on the blanket, but which can identify a consistent $e$ to identify with and thus partake in the same dynamics.

Interestingly, one can also directly compute the dynamics of the sufficient statistic $\boldsymbol{e}$ through the chain rule in terms of the dynamics of the blanket states, which do follow the Ao decomposition. If we do this, we obtain,
\begin{align*}
    \dot{\boldsymbol{e}}(b) &= \frac{\partial \boldsymbol{e}(b)}{\partial b} \frac{db}{dt} \\
    &= \frac{\partial \boldsymbol{e}(b)}{\partial b} (Q_{bb} - \Gamma_{bb}) \nabla_b \ln p^*(i,s,a)
\end{align*}
which is significantly different from the required dynamics as it is a gradient descent on the blanket states $b$ rather than the most likely external state. It is unclear under what conditions we should expect these dynamics to coincide.

\subsubsection{Interpretation of Bayesian Inference Lemma in terms of Average Flows}

An alternative interpretation of the Bayesian inference lemma, which solves some of the issues raised in the previous section, while creating others, is that instead of interpreting the FEP as a statement about the dynamics of the average (since for Gaussian systems the mode and the average coincide) but rather as a statement about the average of the dynamics. On this view, the most likely state does perform a gradient descent on the free energy, but rather the states of the system, on average, perform a gradient descent, or at least that the instantaneous \emph{flow} of the syste is, on average, pointed in the direction of the free energy gradient. This second interpretation is hinted at in \citep{friston2020some} and has been developed more formally in private correspondence, as well as analysed in detail in \citet{aguilera2021particular}. The general result is that while this new interpretation solves the issue of assuming unrealistic dynamics for $\boldsymbol{e}$, it raises two new issues. 

The first, more technical, concerns the $\sigma$ function which must now be redefined to be a mapping between the average flows, rather than between the modes of the internal and external states. Again, this can be achieved in linear Gaussian systems where an analytical solution can be found, but in more complex scenarios the existence and uniqueness of such a mapping cannot be guaranteed. A second, more philosophical issue concerns what this statement about the average dynamics \emph{means}. Specifically, the FEP can no longer be taken to offer even an interpretation of the dynamics of any specific system. It can instead only offer an interpretation of the average behaviour over some counterfactual ensemble of possible systems. Moreover, the average dynamics and the actually realized dynamics can diverge quite strongly in real systems, and can even diverge for Gaussian linear systems, as demonstrated by \citep{aguilera2021particular}. Sometimes, even if a theory only deals in statistical averages, it can be highly fruitful scientifically. For instance, the theory of evolution only ever makes statements about the average changes in populations, not about the behaviour of any specific individual. However, it is unclear at present whether the revised FEPs statement that, on average, the dynamics of a system tend towards a solenoidal gradient flow on the NESS density for that system and thus, on average, tend to minimize variational free energy, is similarly scientifically fruitful.

\subsubsection{Potential and Optimal $\xi$ Functions}

A further interesting question concerns the degree to which it is necessary to define the sufficient statistics $\boldsymbol{e}$ and $\boldsymbol{i}$ through the argmax over the conditional distribution over the external or internal states given the blanket. While the assumption that the dynamics of $\boldsymbol{e}$ equal the dyanmics of $e$ may impose some constraints for this function for $\boldsymbol{e}$, there are no such constraints in the definition of $\boldsymbol{i}(b)$, so we could, in theory use an arbitrary function $\boldsymbol{i}(b) = \xi(b)$ instead of the argmax. Indeed, we might desire to make this function contain \emph{as much information as possible} about the true conditional distribution of the internal states given the external states, so that when the $\sigma$ function maps this to the sufficient statistic of the external density it can be seen as performing inference with the most information possible between the external and internal states. An additional benefit of defining an arbitrary function for $\xi$ instead of using $\xi(b) = argmax \, p(i| b)$ is that we can make $\xi$ differentiable, which alleviates much of the difficulty of making $\sigma$ differentiable as well.

While this approach brings many benefits, it also has the drawback of the necessity to choose a suitable function $\xi$ which introduces another degree of freedom into the modelling process. One possible condition is that we could chose the optimal $\xi$ to be the one that contains the most information about the internal state or, alternatively minimizes the KL between the approximate conditional distribution over the internal states parametrized vy $\xi$ and the true conditional over the blanket states. That is, we could define,
\begin{align*}
    \xi^* = \underset{\xi}{argmin} \, \KL[q(i; \xi(b)) || p(i | b)]
\end{align*}

This would reduce the number of degrees of freedom of $\xi$ and provide a valid modelling target, although the actual computability of this minimization process is potentially a problem, as is whether this objective is actually optimal. Nevertheless, the use of an arbitrary $\xi$ function for the sufficient statistics of the internal states may yet resolve or ameliorate some of the difficulties with the free energy lemma, and is an interesting inroad to begin understanding various relaxations or extensions to the current incarnation of the free energy principle.


\subsubsection{FEP as a locally valid theory}

Overall, we have seen that the derivation of the FEP requires many fairly restrictive assumptions -- first in the formulation of the kinds of systems that the FEP applies to (autonomous Langevin equations with state-independent diagonal additive white Gaussian noise), and secondly in the formulation of the free energy lemma (no solenoidal coupling between subsets, existence and differentiability of the $\sigma$ function, the dynamics of $\boldsymbol{e}$ being the same as the dynamics of $e$), as well as 

Beyond the purely technical considerations outlined above, there are also crucial, and more intuitive issues which  arise in the modelling of biological systems or indeed complex self-organizing systems are the assumptions of ergodicity and the NESS density including the environmental states. Both of these assumptions, when taken literally, are almost always false. Most interesting systems which self-organize are, almost tautologically, non-ergodic, in that they exhibit a high degree of path-dependence and never come close to exploring their full state space. Secondly, the requirement that all subsets of the markov blanket and the system be at non-equilibrium steady state requires that the \emph{external states}, which are typically taken to be the \emph{environment} also be at steady state for the FEP to apply. This condition intuitively does not intuitively hold for general cases of biotic self organization. For instance, we typically consider the system as maintaining itself in steady state against environmental fluctuations, not that the environment itself is defined to be in steady state with us. In the most obvious case where we define the environment to be the whole universe outside of the system, this is definitely false. If we take a more local approach and define the environment to be some small `bubble' around the system, and then simply model the rest of the universe as Gaussian fluctuations impinging on this bubble, then this assumption may be tenable in some cases, but it nevertheless fails to match our intuitions about real biological systems -- such as animals. For instance, it may be the case that my (human) body largely maintains its own homeostatic steady state against external fluctuations, however it is definitely not the case that my environment itself is at steady state at all times -- for instance I can get up and go for a walk, or fly around the world for a conference -- and none of this should disrupt the internal homeostatic steady state of my body. This is the real question which is what the FEP tries (at least in its intuitive sales pitch) to answer -- how can I maintain an internal steady state against an environment which is \emph{not} at steady state. By assuming that the external states are also at the steady state, it may be that the FEP is, in some sense, answering the wrong question and is, in the process, assuming away the true difficulty in answering the right one.

In general, one response to these general issues is to argue that the assumptions of the FEP do not need to hold globally only locally over some relevant timescale, which is a fair point. This idea of the FEP as a theory which only holds locally is valid (although it still needs to be empirically or mathematically shown whether such assumptions such as environmental steady state and ergodicity \emph{do} in fact hold locally over relevant timescales), and has close analogies in mathematics and physics where, for instance, linearity assumptions are often used which are only actually true in the infintesimal limit. In general, it is likely that, in fact, the FEP is, in this sense, a local theory, and should be thought of as such. This locality should then inform modelling work, as well as spur new theoretical advances as to the range of conditions over which the FEP is valid. Understanding and precisely quantifying the limitations and region of applicability of the theory is ultimately vital for obtaining true understanding, and is a very important area for future empirical work within the FEP paradigm.

\section{Mathematical Appendices}
\subsection{Helmholtz Decomposition at Steady State}
\label{helmholtz_decomp_appendix}

It is straightforward to verify that the Helmholtz decomposition of the dynamics satisfies the steady state condition $\frac{dp^*(X)}{dt} = 0$ by plugging this form into the Fokker-Planck equation (Equation \ref{Fokker_Planck}),
\begin{align*}
\frac{dp^*(x)}{dt} &= - \nabla_x \big[(\Gamma - Q)\nabla_x \ln p^*(x) \big] p^*(x) + \Gamma \nabla^2_x p^*(x) \\
 &= - \nabla_x \big[(\Gamma - Q)\frac{\nabla_x p^*(x)}{p^*(x)} \big] p^*(x) + \Gamma \nabla^2_x p^*(x) \\
&= - \nabla_x \big[(\Gamma - Q) \nabla_x p^*(x) \big] + \Gamma \nabla^2_x p^*(x) \\
&= -\Gamma \nabla^2_x p^*(x) + \nabla_x Q \nabla_x p^*(x) + \Gamma \nabla^2_x p^*(x) \\
&= \nabla_x Q \nabla_x p^*(x) = 0 \numberthis
\end{align*}

Where the last line follows because, by definition, the gradient of the solenoidal flow with respect to the gradient of the log density is 0, since the solenoidal flow must be orthogonal to the gradient of the density, which is represented by the solenoidal $Q$ matrix being antisymmetric $Q = -Q^T$.




\end{document}